%% file: paper.tex
\documentclass{article} 
\usepackage{paper,times}
\usepackage{hyperref}
\usepackage{url}
\usepackage{rotating}
\input{vcl-shortcuts.tex}

\title{Conformation Generation using Transformer Flows}

\author{\And Sohil Atul Shah\thanks{This paper was completed in December 2022.}\\
Intel Labs\\
\And 
Vladlen Koltun\footnotemark[1]\\
Intel Labs\\
}

\iclrfinalcopy 
\begin{document}

\maketitle

\begin{abstract}
Estimating three-dimensional conformations of a molecular graph allows insight into the molecule's biological and chemical functions. Fast generation of valid conformations is thus central to molecular modeling. Recent advances in graph-based deep networks have accelerated conformation generation from hours to seconds. However, current network architectures do not scale well to large molecules. Here we present ConfFlow, a flow-based model for conformation generation based on transformer networks.
In contrast with existing approaches, ConfFlow directly samples in the coordinate space without enforcing any explicit physical constraints. The generative procedure is highly interpretable and is akin to force field updates in molecular dynamics simulation. When applied to the generation of large molecule conformations, ConfFlow improve accuracy by up to $40\%$ relative to state-of-the-art learning-based methods. The source code is made available at \href{https://github.com/IntelLabs/ConfFlow}{https://github.com/IntelLabs/ConfFlow}.
\end{abstract}

\section{Introduction}

A fundamental challenge in chemistry and materials science is to deduce valid and stable 3D geometric structures of candidate molecules. 
These structures, usually represented by a set of atomic coordinates, are used in tasks such as molecular property prediction \cite{gilmer2017neural, gebauer2019symmetry}, molecular dynamics (MD) simulation, docking \cite{meng2011molecular, ferreira2015molecular}, and structure-based virtual screening~\cite{cheng2012structure}. Traditionally, x-ray crystallography and density functional theory (DFT) computation played a central role in computing molecular conformations. However, these are computationally expensive and prohibitively time-consuming for industrial-scale prediction on large molecules. Fast refinement of large molecule structures can be realized by applying classical force fields, such as UFF \cite{rappe1992uff} and MMFF \cite{halgren1996merck}, but these suffer from limited accuracy.

High-quality reference 3D structures for drug-like molecules have recently become available \cite{axelrod2020geom}. This opens the possibility of training generative models on large datasets to predict distributions of 3D conformations. Indeed, deep learning has been used in many recent approaches to fast conformation generation. An important focus in this line of work is to ensure equivariance under translation and rotation. A common approach is to embed physical symmetries by structuring the network such that it operates on interatomic pairwise distances, which possess the desired invariances. Given any molecular graph, the network first estimates interatomic distances \cite{Simm2020GraphDG, xu2021end, xu2021learning} or their gradient fields \cite{shi*2021confgf, luo2021predicting}, which are then processed by a heuristic system that satisfies distance geometry, thus producing the final 3D atomic coordinates. Although such models maintain symmetries by construction, the generated distances may be inconsistent: for example, they may violate the triangle inequality. These inconsistencies may propagate into 3D coordinates, yielding inaccurate conformations.

A straightforward alternative is to directly predict 3D coordinates, without first synthesizing interatomic distances or gradients~\cite{mansimov2019molecular,satorras2021}. An early attempt used a conditional variational graph autoencoder (CVGAE) \cite{mansimov2019molecular}, while later work enforced symmetries by applying a rotation equivariant graph neural network (EGNN) \cite{satorras2021}.
Unfortunately, the accuracy of these approaches has remained well below distance-based schemes.

\begin{figure*}[tbhp]
\centering
\includegraphics[width=\linewidth]{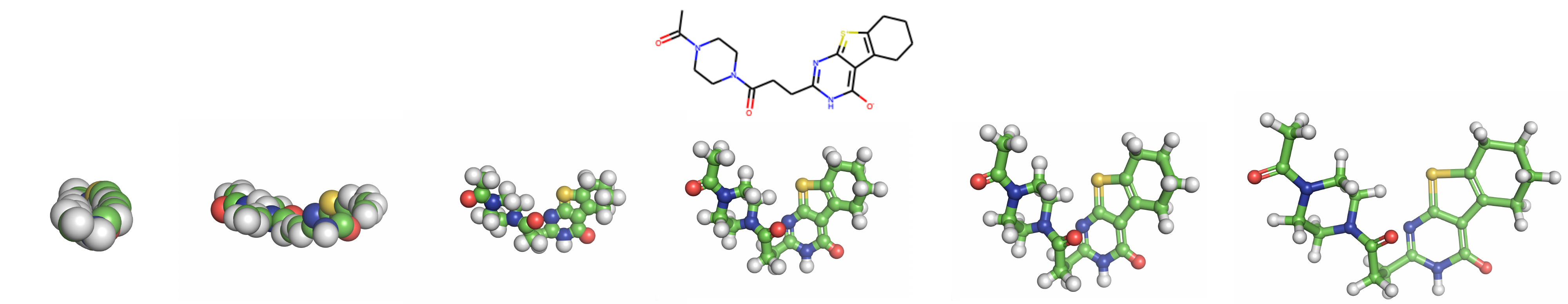}
\caption{ConfFlow transforms points sampled from a simple prior to a 3D molecular conformation. Videos are provided in the supplement.}
\label{fig:sequence}
\end{figure*}

Recent work on related problems has achieved impressive results while operating directly on atomic coordinates. A direct method for protein structure prediction has shown remarkable accuracy \cite{jumper2021highly} and recent work on molecular property prediction has demonstrated promising results without enforcing rotation equivariance \cite{hu2021forcenet,godwin2021very}.
Motivated by these works, we develop a direct method for highly accurate prediction of 3D atomic structure. We build upon the point transformer architecture for 3D point cloud processing \cite{zhao2021point} to design a scalable message passing neural network (MPNN) \cite{gilmer2017neural} that encodes intermediate conformations with no constraints. We integrate the MPNN within a continuous normalizing flow (CNF) framework that performs iterative refinement of atomic coordinates.
The generation process, illustrated in Fig.~\ref{fig:sequence}, can be viewed as analogous to molecular dynamics (MD) simulation where starting from an initial conformation each refinement step gradually moves atoms until they converge to an equilibrium. Our model does not enforce any physical constraints. Rather, we deal with translation invariance by adding a normalization layer at the end of each CNF block.

We conduct extensive experiments on both large and small molecules in the GEOM benchmark \cite{axelrod2020geom}, which contains 250K molecular conformation pairs of Drugs and QM9 molecules. We compare our model, ConfFlow, against state-of-the-art learning-based approaches on multiple tasks. ConfFlow improves over baselines by up to $40\%$ across multiple metrics of conformation generation accuracy.

\section{Formulation}
Consider a set of molecules  $\M = \{\M_k\}_{k=1}^N$. Let each molecule $\M_k$ be represented by a 2D planar embedding expressed using an undirected graph $\G_k = (\V_k, \E_k)$. $\V_k = \{v_i\}_{i=1}^M$ is a set of nodes representing atoms and $\E_k$ is a set of edges signifying interaction between a pair of atoms. Each atom $v_i$ is characterized by a vector of atomic attributes $\hh_i \in \Re^a$ while an edge $e_{ij}$ connecting the $i^{\text{th}}$ and $j^{\text{th}}$ atom is defined by an edge attribute vector $\hh_{ij}\in \Re^b$. The set $\E_k$ represents all chemical bonds between atoms. Since edges between bonded atoms are not sufficient to completely characterize complex atomic interaction in a molecule, we follow usual practice \cite{Simm2020GraphDG} and extend $\E_k$ to include auxiliary edges. These edges are drawn between atoms that are second- or third-level neighbors. The edges between second-level neighbors aid in fixing angles between atoms while those between third-level neighbors fix dihedral angles.

Let the atomic coordinates of the $i^{\text{th}}$ atom, $v_i$, be denoted by $\xx_i \in \Re^3$ and let any conformation of a molecule $\M_k$ be represented by a stacked matrix $\XX_k \in \Re^{M\timess 3}$. 
Based on the atomic stability and external factors, molecules continuously transform between different conformations at equilibrium and can thus have multiple valid conformations.
Given a molecular graph $\G_k$, the task of molecular conformation generation is to generate a set of possible conformations for the molecule. 
To this end, we want to build a generative model that learns the conditional distribution over all possible conformations $p(\XX_k | \G_k)$ and allows us to efficiently sample from this distribution given a molecular graph $\G$.

Formally, we aim to solve the following optimization problem:
\begin{equation}
\F_\btheta\paren{\M} = \argmax_{\F_\btheta} \frac{1}{\abs{\S}} \sum_{k=1}^{\abs{\S}} \log P_{\F_\btheta}\paren{\XX_k | \G_k} \label{eq:opt}
\end{equation}
where $\F_\theta$ is a generative model parameterized by $\theta$, and $\S$ is a set of pairs $\{\paren{\G_k, \XX_k}\}_{k=1}^{\abs{\S}}$ that each provide a molecular graph and a corresponding conformation. We assume that multiple conformers of each molecule are represented as distinct datapoints.

\subsection{Graph-conditioned Normalizing Flows}
We propose to use normalizing flows to model the distribution of conformations for a given molecular graph $\G$.
A normalizing flow is a generative model that defines complex distribution using a series of invertible bijective transformations of random variables over a simple base distribution \cite{rezende2015variational, dinh2016density}. Since it is invertible, any data point from an induced distribution can be mapped back into the base distribution using a change of variables, which enables exact likelihood computation.
Consider a learnable function with parameters $\btheta$, $\F_\btheta:\Re^3 \to \Re^3$, given by a series of invertible and differentiable transformations, $\F_\btheta = f_L \circ f_{L-1} \dots \circ f_1$, and a random variable $\YY \in \Re^{M\times 3}$ sampled from a base density distribution $P_y$. The exact log-likelihood of output variable $\XX = \F_\btheta\paren{\YY}$ under the distribution $P_x$ is given by
\begin{equation}
\log P_x\paren{\XX} = \log P_y\paren{\YY} - \sum_{l=1}^L \log \abs{\det \frac{\partial f_l}{\partial f_{l-1}}} \label{eq:nf}
\end{equation}
where $f_0 = \YY$ and $\frac{\partial f_l}{\partial f_{l-1}}$ is the Jacobian of function $f_l$. For any given sample $\XX$, one estimates probability density $P_x(\XX)$ by deducing $\YY$ using the inverse transformation $\YY = \F_\btheta^{-1}(\XX)$ and applying \eqref{eq:nf}.

Normalizing flows are trained by maximizing the exact log-likelihood in \eqref{eq:nf}, which is more stable than optimizing over lower bounds of log-likelihood in variational autoencoders (VAEs).
In practice, each $f_l$ is parameterized using a neural network with a simple Jacobian (such as a lower triangular matrix) whose determinant is computationally tractable.
Since our goal is to learn a flow model for $\XX \sim P_x(\XX | \G)$ that maps atoms in a given 2D molecular graph to its 3D position, we characterize $f_l$ using a reversible graph processing layer. These layers are then interleaved with graph-independent reversible batch normalization layers. 

For graph processing layer one may use any variant of MPNN \cite{gilmer2017neural}, a widely used framework for representation learning on molecular graphs. However, the challenge is in repurposing these layers to be invertible while preserving expressiveness. We choose to characterize each graph processing layer using a CNF \cite{chen2018neural, grathwohl2019ffjord}, which is readily invertible and supports unrestricted transformation and interaction between atomic coordinates in the latent space.

The CNF architecture generalizes normalizing flows from a discrete number of layers to a continuous flow with an invertible transformation $f_l$ defined using a differential equation $f_l\paren{\ZZ^l(t), t; \G} = \frac{\partial \ZZ^l(t)}{\partial t}$, while the change in log-density is given by $\frac{\partial \log p(\ZZ^l(t))}{\partial t} = \trace \paren{\frac{\partial f_l}{\partial \ZZ^l(t)}}$. The output variable $\XX^l$ and its exact log-likelihood are now obtained by integrating across time:
\begin{subequations}
\label{eq:cnf}
\begin{align}
&\XX^l = \ZZ^l(t_0) + \int_{t_0}^{t_1}  f_l\paren{\ZZ^l(t), t; \G} dt \text{  s.t.  } \ZZ^l(t_0) = \XX^{l-1} \\
&\log P_x(\XX^l | \G) = \log P_z(\ZZ^l(t_0)) - \int_{t_0}^{t_1} \trace \paren{\frac{\partial f_l}{\partial \ZZ^l(t)}} dt
\end{align}
\end{subequations}
while an inverse transformation is realized by reversing the order of integration: $\ZZ^l(t_0) = \XX^l + \int_{t_1}^{t_0}  f_l\paren{\ZZ^l(t), t; \G} dt$, where $\ZZ^l(t_1) = \XX^l$. In practice, the trace is approximated using the Hutchinson trace estimator \cite{hutchinson1989stochastic}, while the integral and backpropagation are straightforwardly computed using a black-box ordinary differential equation (ODE) solver \cite{chen2018neural}. In comparison to neural layers in discrete flows, only mild constraints are enforced on functions $f_l$ -- second-order differentiability and Lipschitz continuity, with a possibly large Lipschitz constant, which is readily satisfied by using smooth activations.

\begin{figure*}[tbhp]
\centering
\includegraphics[width=1.0\linewidth]{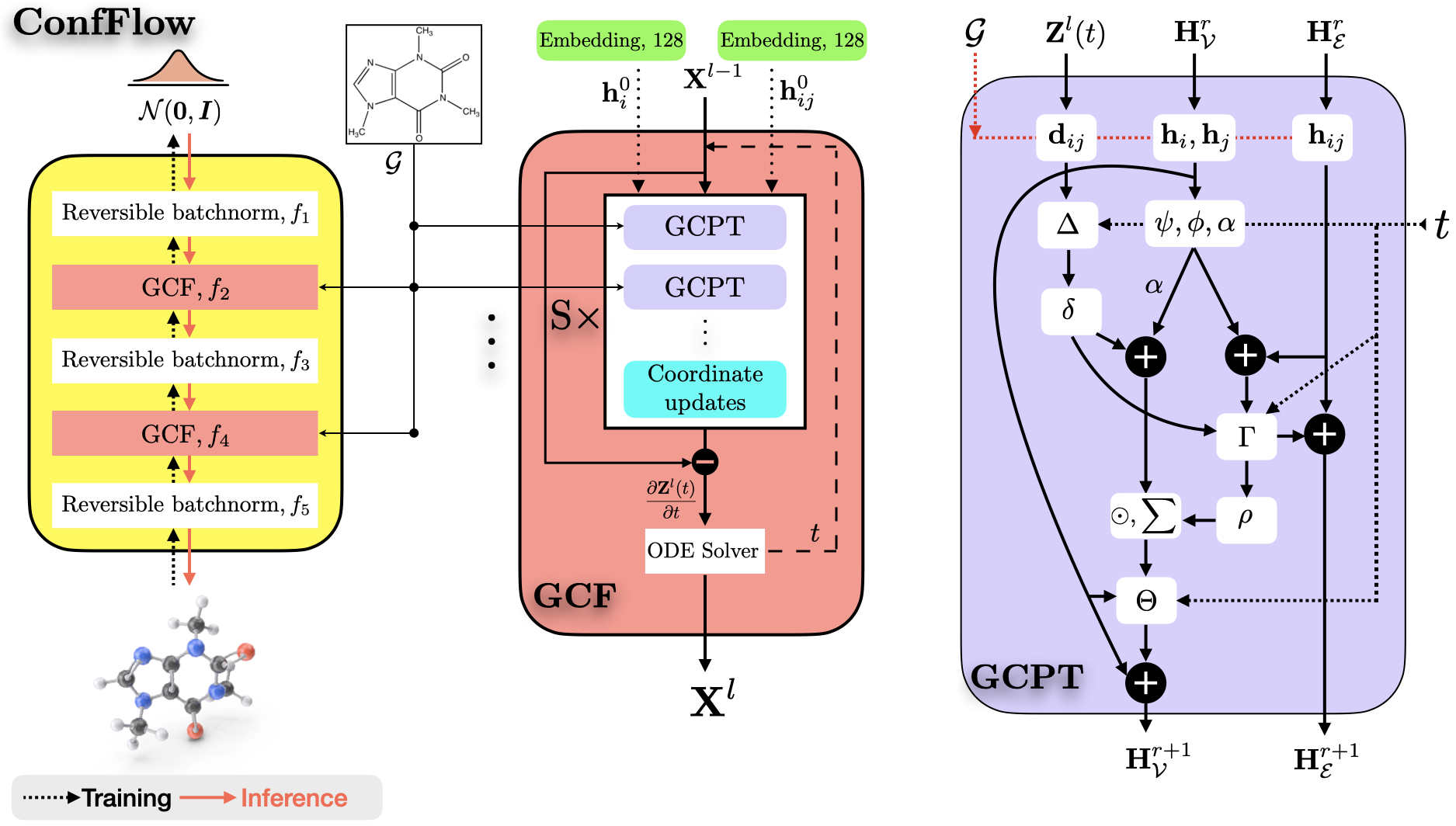}
\caption{Illustration of ConfFlow with an architectural overview (left), a graph continuous flow (GCF) block (middle), and a graph conditional point transformer (GCPT) message passing layer used within GCF (right).}
\label{fig:gcnf}
\end{figure*}

\subsection{GCF: Graph Continuous Flows}
We implement the dynamics $f_l$ of the continuous flow using an MPNN based on point transformer~\cite{zhao2021point}, a self-attention layer for point cloud processing. Fig.~\ref{fig:gcnf} depicts the overall architecture of our graph-conditioned normalizing flow (ConfFlow). Given atom and bond attributes as inputs, they are first embedded into feature space $\Re^e$ using feedforward neural networks:
\begin{align}
\hh^0_i &= \text{NodeEmbedding}(\hh_i), \,\,\,\, i \in \V \\
\hh^0_{ij} &= \text{EdgeEmbedding}(\hh_{ij}), \,\,\,\, ij \in \E
\end{align}
Subsequently, the embeddings, coordinates, and input graph structure $\G$ are processed using $S$ modules of Point Transformer blocks, each containing $R$ message passing layers of graph conditional point transformer (GCPT) and a coordinate update layer.

At layer $r$ and time $t$, GCPT takes as input the set of node embeddings $\HH^r_\V = \{\hh^r_1, \dots,\hh^r_M \}$, edge embeddings $\HH^r_\E = \{\hh_{ij}^r, \, \forall \,\, ij \in \E\}$, and coordinate embeddings $\ZZ^{l}(t) = \{\zz^l_1, \dots, \zz^l_M\}$, and computes a transformation $\HH^{r+1}_\V$ and $\HH^{r+1}_\E$ by aggregating information from neighboring nodes and edges:
\begin{align}
&\dd_{ij} = \zz^l_i - \zz^l_j \,\,\,\, \text{and} \,\,\,\,
\delta = \Delta\paren{\dd_{ij}, t} \label{eq:posenc} \\
&\hat{\hh}_{ij}^{r+1} = \Gamma \paren{\psi\paren{\hh^r_i, t} - \phi\paren{\hh^r_j, t} + \hh_{ij}^r, \delta, t} \label{eq:mes} \\
&\mm_i^{r+1} = \sum_{j \in \N_i} \rho\paren{\hat{\hh}_{ij}^{r+1}} \odot \paren{\alpha\paren{\hh_j^r, t} + \delta} \label{eq:mesagg} \\
&\hh_i^{r+1} = \hh_i^{r} + \Theta\paren{\hh_i^r, \mm_i^{r+1}, t} \label{eq:noderes} \\
&\hh_{ij}^{r+1} = \hh_{ij}^{r} + \hat{\hh}_{ij}^{r+1} \label{eq:edgeres}
\end{align}
Here $\N_i$ is a set of neighbors connected to the $i^{\text{th}}$ node in graph $\G$ and $\rho:\Re^e \to [0,1]^e$ is a softmax normalization that functions as a vector attention mechanism over neighboring nodes. $\phi, \psi, \alpha$ and $\Delta, \Gamma, \Theta$ denote time-dependent linear functions and multi-layer perceptrons (MLPs), respectively. Multiple input variables within $\paren{.}$ are concatenated along the feature dimension.
$\Delta$ in \eqref{eq:posenc} represents a position encoding function whereas \eqref{eq:mes} and \eqref{eq:mesagg} resemble message computation and message aggregation operations of an MPNN, respectively.

Coming back to the point transformer block in GCF, note that the coordinate embeddings are shared across $R$ message passing layers while the node and edge features are regularly updated using residual connections in \eqref{eq:noderes} and \eqref{eq:edgeres}. After $R$ rounds of message passing using GCPT, these transformed embeddings are fed into the MLP layer, $\Omega$, which independently updates each coordinate embedding using
\begin{align}
\tilde{\zz}_i^l &= \Omega\paren{\zz_i^l, \hh_i^{R}, \frac{1}{\abs{\N_i}}\sum_{j \in \N_i} \hh_{ij}^{R}, t}
\end{align}
while combining the node and its corresponding edge features. As they are processed using $\Omega$ at each of the $S$ modules of Point Transformer, these input embeddings at layer $f_l$, $\zz^{l-1}\in\Re^3$,  are transformed to higher dimensionality. The final $\Omega$ layer projects these embeddings back to 3D Euclidean space following which the dynamics of $f_l$ at time $t$ are given by
\begin{align}
\frac{\partial \ZZ^l(t)}{\partial t} = \paren{\frac{\partial \zz^l_1}{\partial t}, \dots, \frac{\partial \zz^l_M}{\partial t}} \,\,\,\, \text{where } \,\,
\frac{\partial \zz^l_i}{\partial t} = \tilde{\zz}_i^l - \zz_i^l
\end{align}
As $t \to \infty$, the continuous-time dynamics in neural ODEs can be viewed as unrolled infinite-layer residual networks~\cite{chen2018neural}. Similarly, GCF unrolls into infinite message passing layers and is thus capable of modeling long-range dependencies.

\subsection{Regularization}
The computational cost of numerically integrating ODEs is prohibitively high.
Moreover, while training on large molecules we find the dynamics of ODE solver to be numerically unstable.
In order to effectively learn $f_l$ we regularize the dynamics with two terms \cite{finlay2020train}, the kinetic energy of the flow and the Jacobian Frobenius norm:
\begin{subequations}
\label{eq:reg}
\begin{align}
\F^K_l &= \int_{t_0}^{t_1} \norm{f_l\paren{\ZZ^l(t), t; \G}}^2 dt \label{eq:reg1}\\
\F^J_l &= \int_{t_0}^{t_1} \norm{\nabla_{\ZZ}f_l\paren{\ZZ^l(t), t; \G}}^2_F dt \label{eq:reg2}
\end{align}
\end{subequations}
where \eqref{eq:reg1} penalizes distance travelled by atoms under the flow $f_l$ and \eqref{eq:reg2} improves generalization by regularizing the Jacobian. In trace form, the Jacobian Frobenius norm is easily approximated using the Hutchinson trace estimator without any additional computational cost.

\subsection{Training Objective}
Combining Eqs.~(\ref{eq:opt}, \ref{eq:nf}, \ref{eq:cnf}, \ref{eq:reg}), the final optimization problem to be solved is
\begin{multline}
\F_\btheta =  \argmax_{\F_\btheta} \frac{1} {3\A} \sum_{k=1}^{\abs{\S}} \left[ \log P_z(\ZZ_k) - \sum_{l = 1,3,..}^L \log \abs{\det \frac{\partial f_l}{\partial f_{l-1}}} 
-\sum_{l=2,4,..}^L \int_{t_0}^{t_1}  \trace \paren{\frac{\partial f_l}{\partial \ZZ_k^l(t)}} \right. \\ 
-\lambda_K \norm{f_l\paren{\ZZ^l_k(t), t; \G_k}}^2 
-  \lambda_J \norm{\nabla_{\ZZ}f_l\paren{\ZZ^l_k(t), t; \G_k}}^2_F \,\, dt \Bigg]
\end{multline}
where $\ZZ_k = f^{-1}_l\paren{\XX_k; \G_k}$ as we transform $\XX_k$ backwards through ConfFlow. Note that instead of total input graph pairs $\abs{\S}$, we average over the total number of atoms $\A=\sum_k^{\abs{\S}} \abs{\V_k}$ and dimension. This ensures that weights $\lambda_K$ and $\lambda_J$ can be fixed independently of the size of molecules across batches and datasets. By simply augmenting the dynamics of the ODE with multiple state vectors, all terms within the integral are numerically integrated using a \textit{single} ODE solver.

\subsection{Sampling}
The invertibility of $\F_\btheta$ enables fast sampling. To generate a molecular conformer for a given graph representation $\G = (\V,\E)$, we first independently sample a set of latent vectors $\ZZ \doteq \left\{\zz_i \in \Re^3, 1 \leq i \leq \abs{V}\right\}$ from standard Gaussian space $\N(0, I)$, mapping each to a corresponding atom and then processing $\ZZ$ via $\F_\btheta$ conditioned on $\G$ to produce 3D coordinates of atoms in a molecule, $\XX = \F_\btheta\paren{\ZZ; \G} \in \Re^{\abs{V}\times 3}$.

\subsection{Implementation}
Our model is implemented in PyTorch~\cite{paszke2019pytorch}. We use the same settings across all datasets. Our ConfFlow architecture contains two blocks of GCF each stacked between reversible batchnorm layers as shown in Fig. \ref{fig:gcnf} (left). GCF employs $S=3$ point transformer blocks with output embedding dimensionality fixed to $32$, $32$, and $3$, respectively. Each point transformer block contains $R=2$ GCPT message passing layers. All the MLPs within GCPT are implemented with two linear layers and a Swish nonlinearity \cite{ramachandran2017searching, elfwing2018sigmoid}. We solve the dynamics for $\eqref{eq:cnf}$ using a Runge-Kutta 4(5) adaptive solver~\cite{dormand1980family} with error tolerances $1e^{-3}$, while the backpropagation through solver steps is realized using a memory efficient adjoint method \cite{farrell2013automated, chen2018neural}. The coefficients of both regularization terms are set to $0.2$. We train ConfFlow with two GPUs for 32K iterations using the Adam optimizer \cite{kingma2015adam} with learning rate $1e^{-3}$, per-GPU batch size $125$, and gradient norm restricted to $0.05$. SI lists the full set of node and edge attributes. We preprocess features by normalizing them across datasets and mapping them into an input embedding space of dimensionality $e=128$.

\section{Experiments}

We assess the performance of ConfFlow by comparing against state-of-the-art learning-based methods for molecular conformation generation. We use the following standard tasks. a) \textbf{Conformation Generation} tests the model's ability to learn the distribution of conformations by measuring the diversity and accuracy of generated samples. b) \textbf{Property Prediction} measures the accuracy of a predicted chemical property for a molecular graph.

\subsection{Data}
Following the protocol in \cite{shi*2021confgf}, we benchmark conformation generation and property prediction on GEOM data \cite{axelrod2020geom}, which provides multiple high-quality stable conformations per molecule. The GEOM-QM9 dataset, an extension of QM9 \cite{ramakrishnan2014quantum}, contains multiple conformations for small molecules with up to nine heavy atoms ($29$ total atoms). The GEOM-Drugs dataset consists predominantly of medium-sized organic compounds having an average of $44$ and a maximum of $181$ atoms per molecule. For each of these datasets we randomly sample 40,000 molecules and select the five most likely conformations per molecule, yielding a training set of 200,000 conformation-molecule pairs \cite{shi*2021confgf}. Another $200$ randomly sampled molecules and their corresponding conformations constitute a test split of 22,408 and 14,324 conformation-molecule pairs, for GEOM-QM9 and GEOM-Drugs respectively.

\subsection{Baselines}
We benchmark ConfFlow against eight state-of-the-art methods for conformation generation. \textbf{CVGAE} \cite{mansimov2019molecular} is a variational graph auto-encoder that processes any given molecular graph through multiple layers of GRUs to directly generate 3D atomic coordinates. \textbf{GraphDG} \cite{Simm2020GraphDG} and \textbf{CGCF}~\cite{xu2021learning} employ conditional graph VAE and CNF networks, respectively, to learn distributions over interatomic distances. These distances are later post-processed using a Euclidean distance geometry (EDG) algorithm to generate 3D conformations. \textbf{ConfVAE} \cite{xu2021end} addresses noise in the sampled distances by training an end-to-end framework using bilevel programming. ConfVAE optimizes for pairwise distances and atomic coordinates jointly. \textbf{ConfGF} \cite{shi*2021confgf}
trains noise-conditional score networks to estimate gradient fields of the log density of atomic coordinates via interatomic distances. Using these estimated gradient fields, the stable conformers are generated using annealed Langevin dynamics.
\textbf{GeoMol} \cite{ganea2021geomol} builds local structure by predicting the coordinates of non-terminal atoms, which are then refined and assembled using torsion angles learned from local distances and dihedral angles. Finally, \textbf{RDKit} \cite{riniker2015better} is a classic EDG approach built upon manually crafted rules for distance-bound matrices. The results for all baselines are obtained by running publicly available code provided by the authors.

\subsection{Conformation Generation}
This task is designed to evaluate the model's capacity to generate realistic and diverse molecular conformations. For each of the unique molecular graphs in the test set, we sample twice the number of conformations found in the reference set. We measure discrepancy between the generated $\RR$ and reference $\RR^{*}$ conformations using Root Mean Squared Deviations (RMSD) of the atoms \cite{mansimov2019molecular, hawkins2017conformation}:
\begin{align}
\text{RMSD}\paren{\RR, \RR^{*}} &= \min_{\AA}\paren{ \frac{1}{n} \sum_{i=1}^n \norm{\RR -  \AA\paren{\RR^{*}}}^2 }^{\frac{1}{2}} \label{eq:rmsd}
\end{align}
Here $n$ denotes the number of atoms in the molecule and $\AA$ is the optimal alignment between two conformations w.r.t.\ rotation and translation. The hydrogen atoms cannot be located accurately using x-ray crystallography \cite{ogata2015hydrogens} and their ground-truth 3D coordinates are therefore contentious. Following standard practice, we restrict evaluation to only heavy atoms unless otherwise stated. Xu et al.~\cite{xu2021learning} define three scores based on RMSD -- Coverage (COV), Matching (MAT), and Mismatch (MIS) -- to measure the diversity, accuracy, and quality of generated samples, respectively:
\begin{align}
\text{COV}\paren{\Se_g, \Se_r} &= \frac{\abs{\left\{ \RR \in \Se_r \,|\, \text{RMSD}\paren{\RR, \hat{\RR}} < \delta, \exists \, \hat{\RR} \in \Se_g \right\}}}{\abs{\Se_r}}  \\
\text{MAT}\paren{\Se_g, \Se_r} &= \frac{1}{\abs{\Se_r}} \sum_{\RR \in \Se_r} \min_{\hat{\RR} \in \Se_g} \text{RMSD}\paren{\RR, \hat{\RR}} \\
\text{MIS}\paren{\Se_g, \Se_r} &= \frac{\abs{\left\{ \RR \in \Se_g \,|\, \text{RMSD}\paren{\RR, \hat{\RR}} > \delta, \forall \, \hat{\RR} \in \Se_r \right\}}}{\abs{\Se_g}} 
\end{align}
Here for any given molecular graph, $\Se_g$ and $\Se_r$ represent generated and reference sets of conformations, respectively. Given an RMSD threshold $\delta$, COV measures the fraction of reference conformations that are matched by at least one of the generated samples. MIS counts the fraction of generated samples that are not covered by any of the reference conformations. MAT focuses on raw accuracy. In general, higher COV corresponds to better diversity while lower MIS and MAT indicate higher accuracy. Following \cite{xu2021learning}, we set the threshold $\delta$ to $0.5$ for GEOM-QM9 and $1.25$ for GEOM-Drugs.

\begin{table*}[htbp]
\centering
\ra{1.05}
\resizebox{\linewidth}{!}{
\begin{tabular}{@{}l@{\hspace{1mm}}|c@{\hspace{2mm}}c@{\hspace{2mm}}c@{\hspace{2mm}}c@{\hspace{2mm}}c@{\hspace{2mm}}c@{\hspace{1mm}}|c@{\hspace{2mm}}c@{\hspace{2mm}}c@{\hspace{2mm}}c@{\hspace{2mm}}c@{\hspace{2mm}}c@{\hspace{1mm}}|c@{}}
\toprule
Dataset & \multicolumn{6}{c|}{GEOM-QM9} & \multicolumn{6}{c|}{GEOM-Drugs} & \\
\midrule
\multirow{2}{*}{Algorithm} & \multicolumn{2}{c}{COV ($\%$)} & \multicolumn{2}{c}{MAT ($\r{A}$)} & \multicolumn{2}{c|}{MIS ($\%$)} & \multicolumn{2}{c}{COV ($\%$)} & \multicolumn{2}{c}{MAT ($\r{A}$)} & \multicolumn{2}{c|}{MIS ($\%$)} & Rank \\
& Mean & Median & Mean & Median & Mean & Median & Mean & Median & Mean & Median & Mean & Median \\
\midrule
CVGAE & 0.14 & 0.00 & 0.867 & 0.863 & 99.60 & 100.00 & 0.00 & 0.00 & 3.070 & 2.994 & 100.00 & 100.00 & 7.7 \\
GraphDG & 14.71 & 2.58 & 0.963 & 0.935 & 78.54 & 98.78 & 2.43 & 0.00 & 2.493 & 2.459 & 97.81 & 100.00 & 7.2 \\
CGCF & 81.11 & 85.90 & 0.393 & 0.363 & 57.74 & 61.00 & 54.50 & 56.00 & 1.245 & 1.227 & 77.80 & 84.50 & 5.0 \\
ConfVAE & 79.11 & 83.40 & 0.405 & 0.384 & 62.32 & 65.40 & 41.40 & 36.10 & 1.332 & 1.335 & 85.80 & 93.40 & 6.0 \\
ConfGF & \underline{90.55} & \underline{95.13} & \textbf{0.269} & \textbf{0.270} & 52.06 & 53.02 & 61.50 & \underline{71.20} & \underline{1.170} & 1.148 & 76.80 & 84.10 & 2.7 \\
GeoMol &  71.18 &  72.76 &  0.375 &  0.375 &\underline{17.84}  & \underline{14.04} & 56.73 & 60.00 & 1.236 & 1.179 & \textbf{22.39} & \textbf{2.90} &  3.0 \\
\textbf{ConfFlow} & \textbf{91.08} &  \textbf{95.60} &  \underline{0.278} &  \underline{0.287} & 47.60 &  50.70 & \textbf{88.30} & \textbf{97.60} & \textbf{0.895} & \textbf{0.874} & 39.80 & 38.30 & \textbf{1.8} \\
\midrule
RDKit & 84.22 & 90.51 & 0.334 & 0.291 & \textbf{8.36} & \textbf{1.31} & \underline{61.60} & 62.40 & 1.195 & \underline{1.139} & \underline{28.90} & \underline{15.10} & \underline{2.5} \\
\bottomrule
\end{tabular}
}
\caption{Comparison of COV, MAT, and MIS scores for different approaches on the GEOM-QM9 and GEOM-Drugs datasets.}
\label{tab:confgen}
\end{table*}
\begin{figure*}[h]
\centering
\ra{1.05}
\resizebox{\linewidth}{!}{
\begin{tabular}{@{}l@{\hspace{1mm}}|c@{\hspace{2mm}}c@{\hspace{2mm}}c@{\hspace{2mm}}|c@{\hspace{2mm}}c@{\hspace{2mm}}c@{\hspace{2mm}}|c@{\hspace{2mm}}c@{\hspace{2mm}}c@{\hspace{2mm}}|c@{\hspace{2mm}}c@{\hspace{2mm}}c@{\hspace{2mm}}c@{}}
\toprule
\rotatebox[origin=l]{90}{Input Graph} & 
\multicolumn{3}{c|}{\includegraphics[height=1.5cm,width=0.245\linewidth]{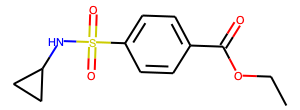}} &
\multicolumn{3}{c|}{\includegraphics[height=1.5cm,width=0.245\linewidth]{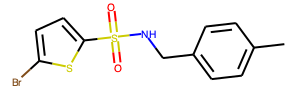}} &
\multicolumn{3}{c|}{\includegraphics[height=1.5cm,width=0.245\linewidth]{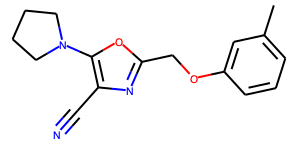}} &
\multicolumn{3}{c}{\includegraphics[height=1.5cm,width=0.245\linewidth]{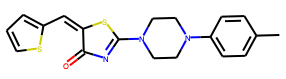}} \\
\midrule
\rotatebox[origin=l]{90}{Ref. Conf.} &
\includegraphics[height=1.4cm,width=0.082\linewidth]{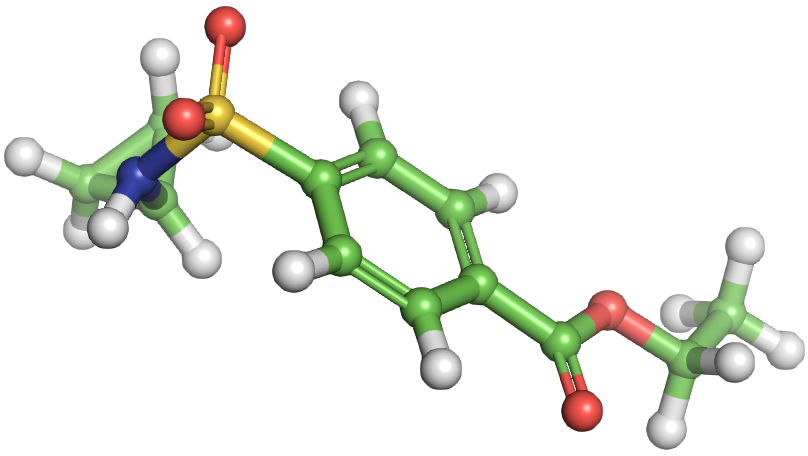} &
\includegraphics[height=1.4cm,width=0.082\linewidth]{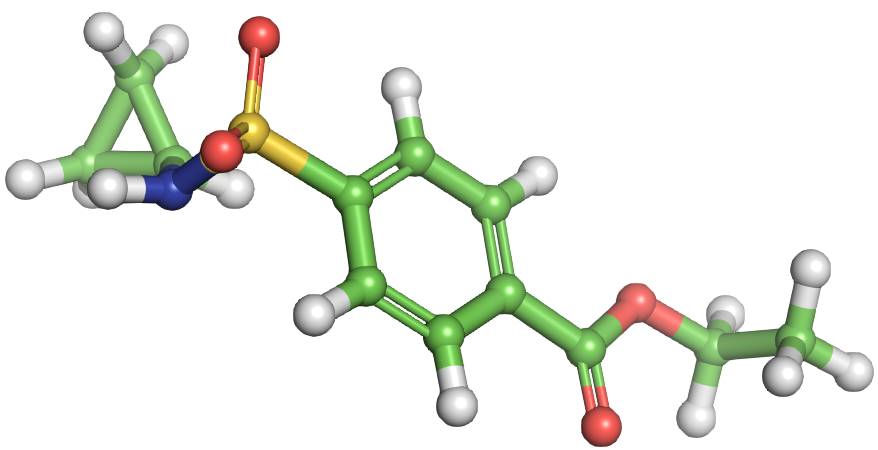} &
\includegraphics[height=1.4cm,width=0.082\linewidth]{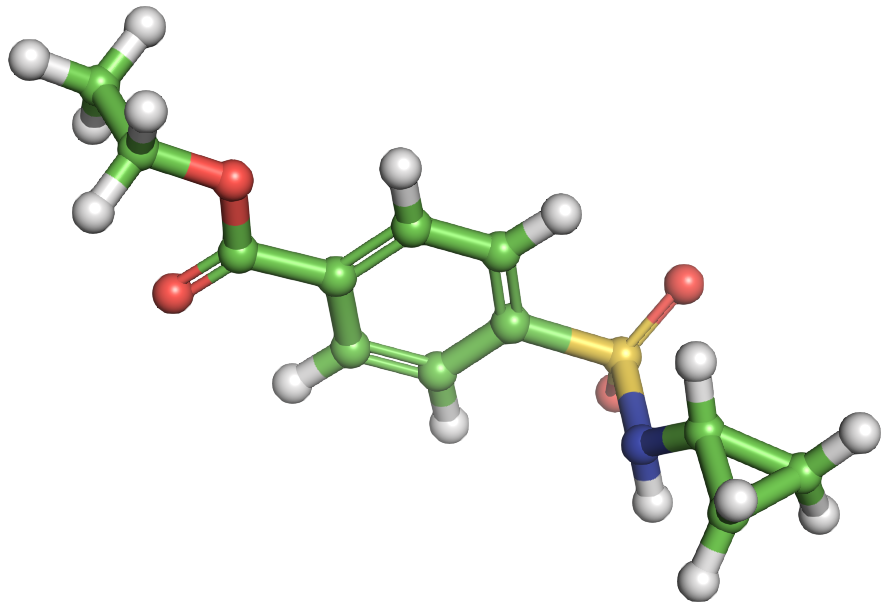} &
\includegraphics[height=1.4cm,width=0.082\linewidth]{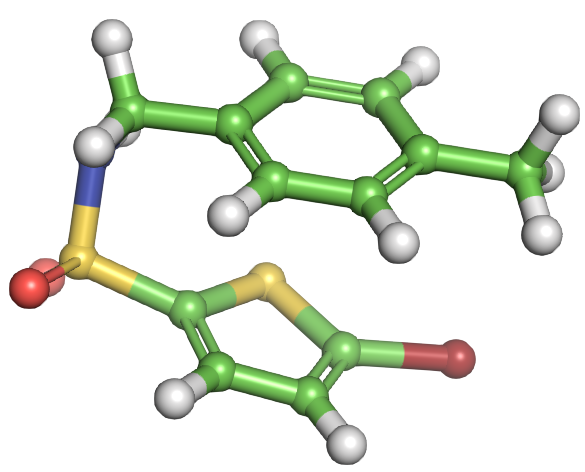} &
\includegraphics[height=1.4cm,width=0.082\linewidth]{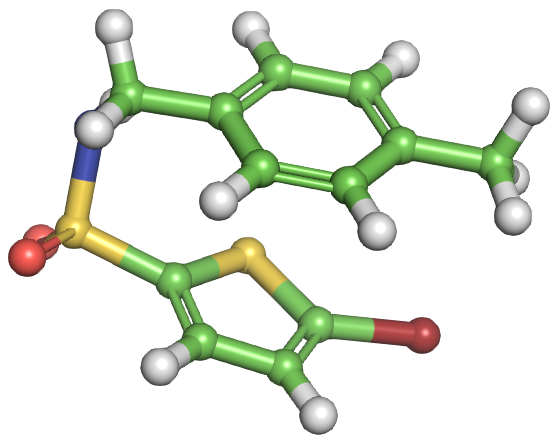} &
\includegraphics[height=1.4cm,width=0.082\linewidth]{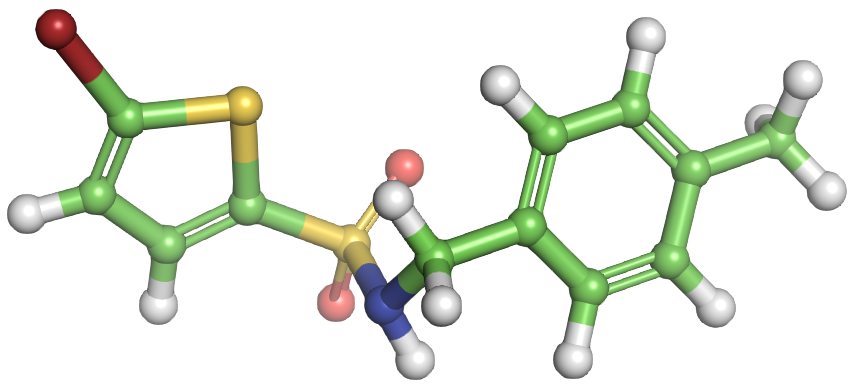} &
\includegraphics[height=1.4cm,width=0.082\linewidth]{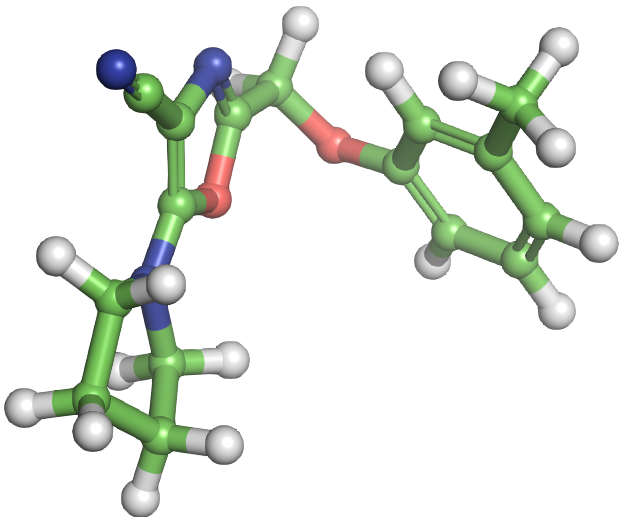} &
\includegraphics[height=1.4cm,width=0.082\linewidth]{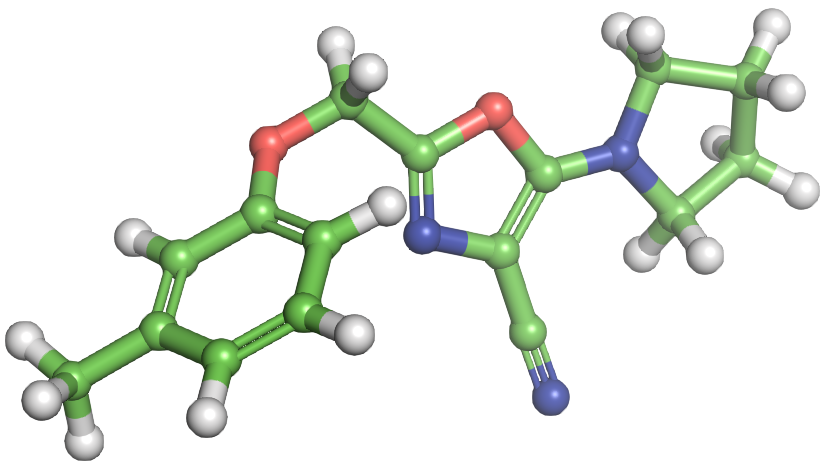} &
\includegraphics[height=1.4cm,width=0.082\linewidth]{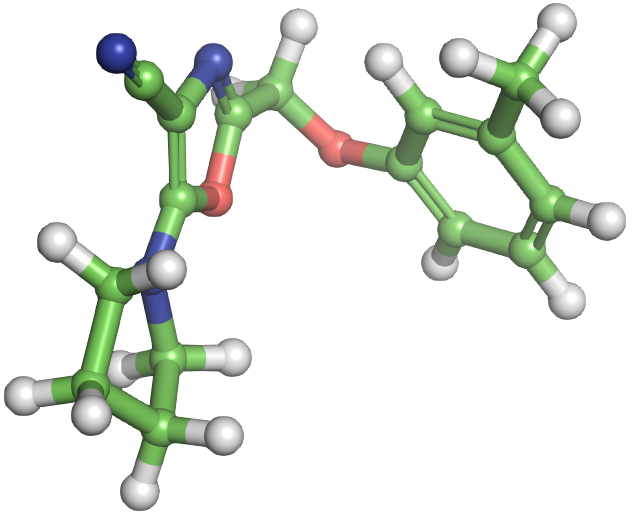} &
\includegraphics[height=1.4cm,width=0.082\linewidth]{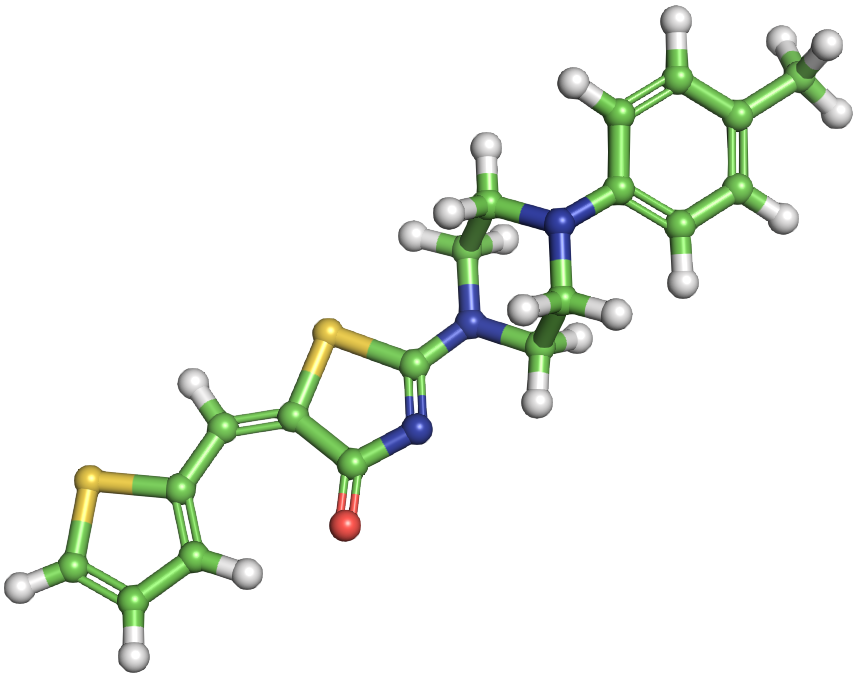} &
\includegraphics[height=1.4cm,width=0.082\linewidth]{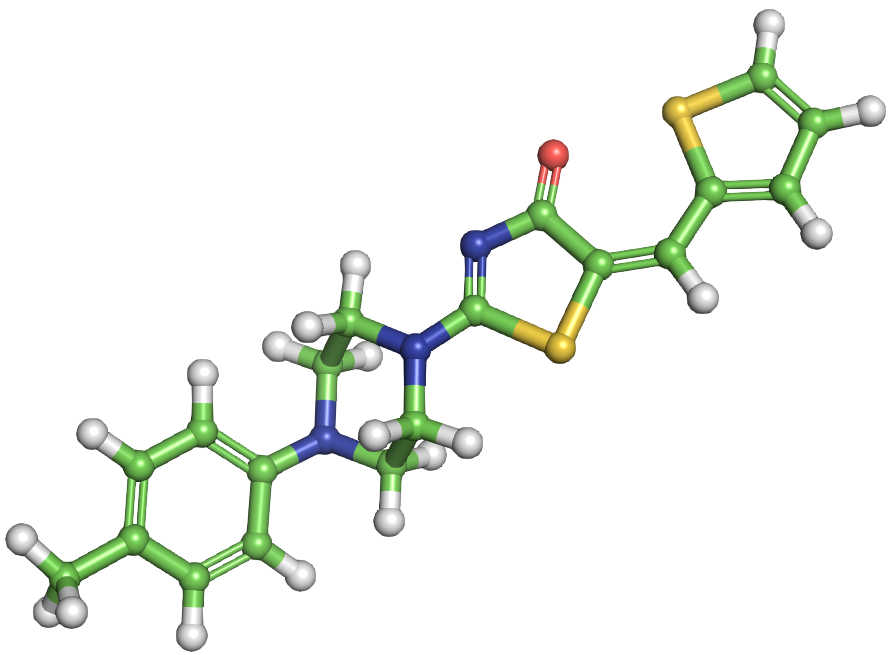} &
\includegraphics[height=1.4cm,width=0.082\linewidth]{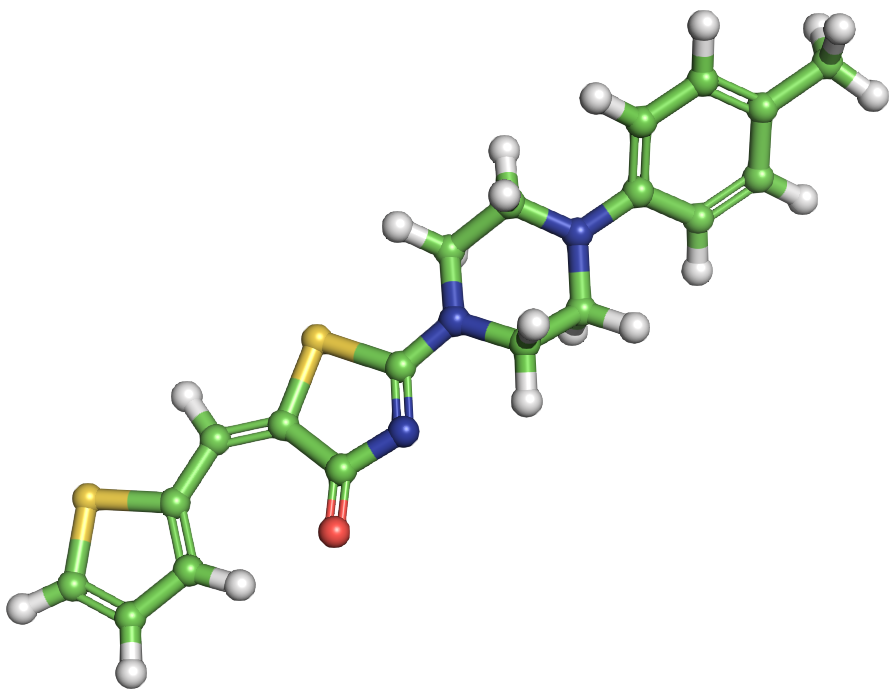}  \\
\rotatebox[origin=l]{90}{\textbf{ConfFlow}} &
\includegraphics[height=1.4cm,width=0.082\linewidth]{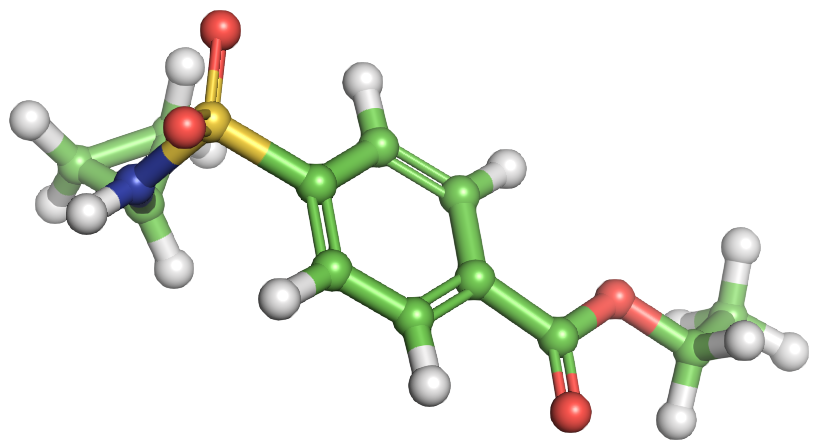} &
\includegraphics[height=1.4cm,width=0.082\linewidth]{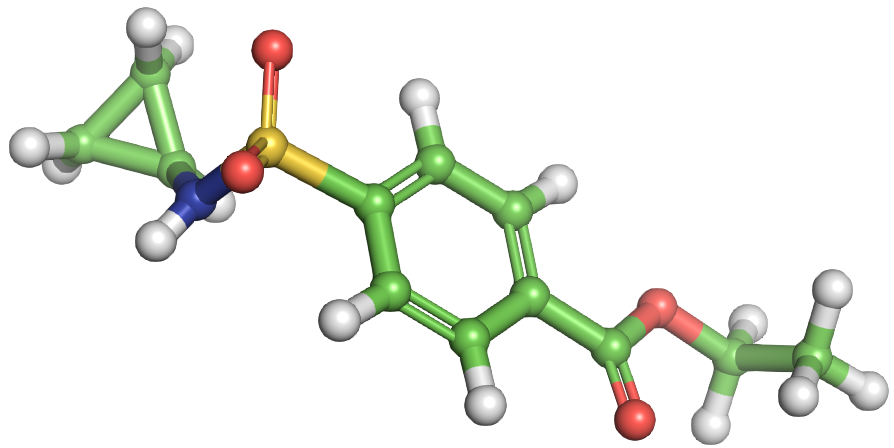} &
\includegraphics[height=1.4cm,width=0.082\linewidth]{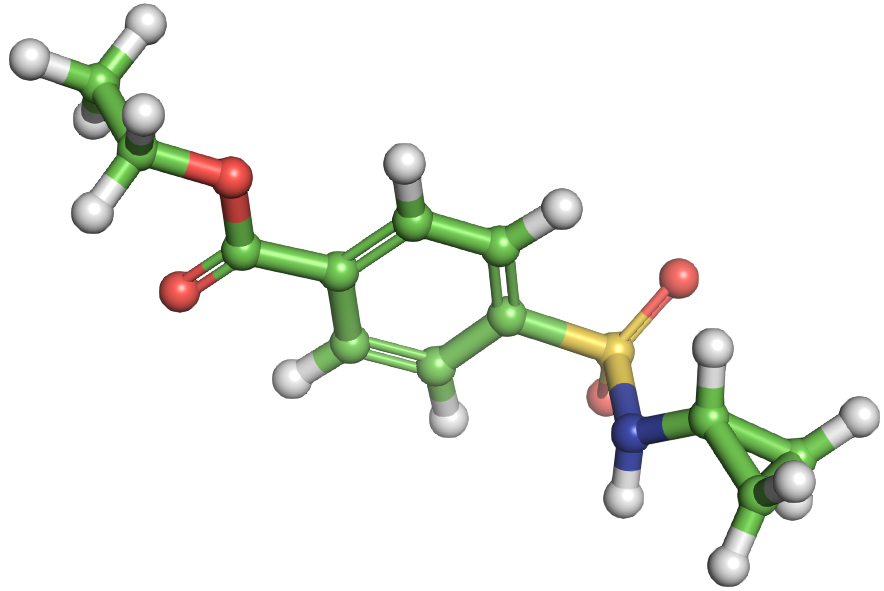} &
\includegraphics[height=1.4cm,width=0.082\linewidth]{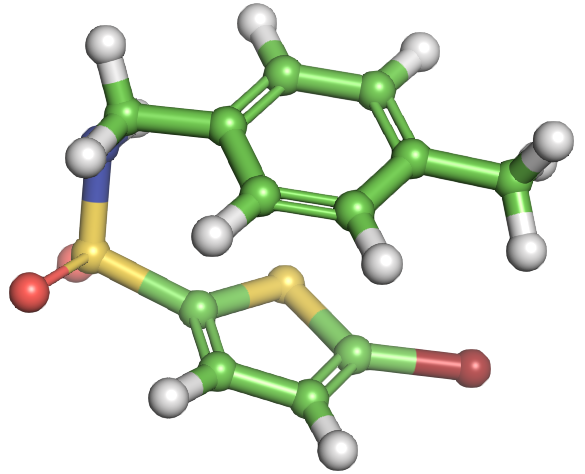} &
\includegraphics[height=1.4cm,width=0.082\linewidth]{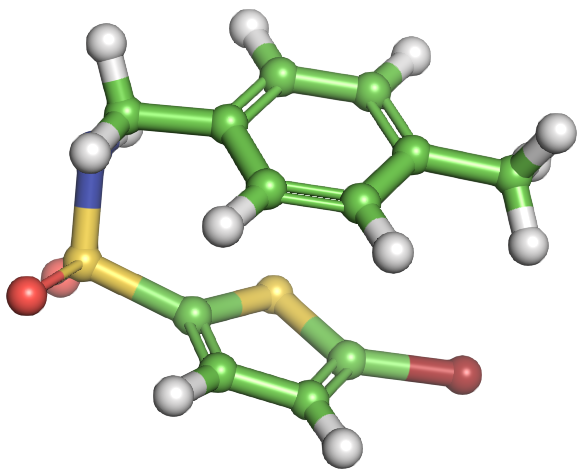} &
\includegraphics[height=1.4cm,width=0.082\linewidth]{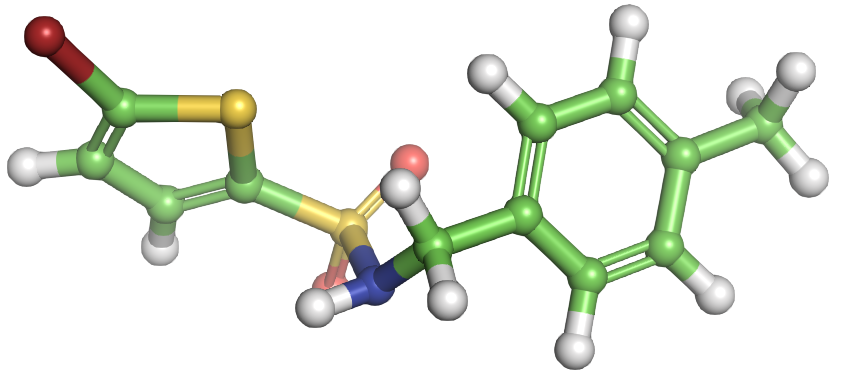} &
\includegraphics[height=1.4cm,width=0.082\linewidth]{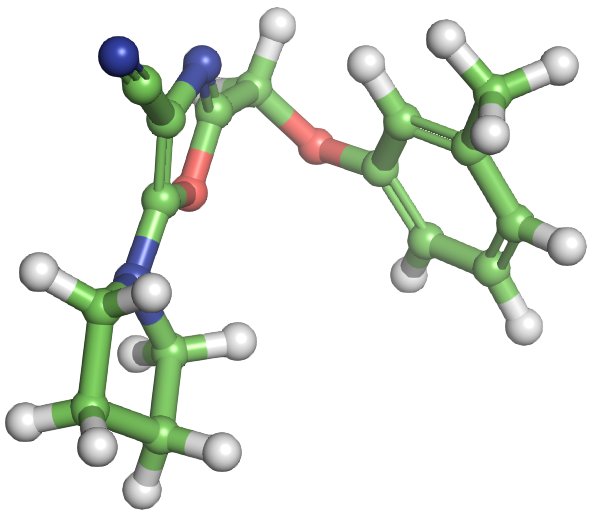} &
\includegraphics[height=1.4cm,width=0.082\linewidth]{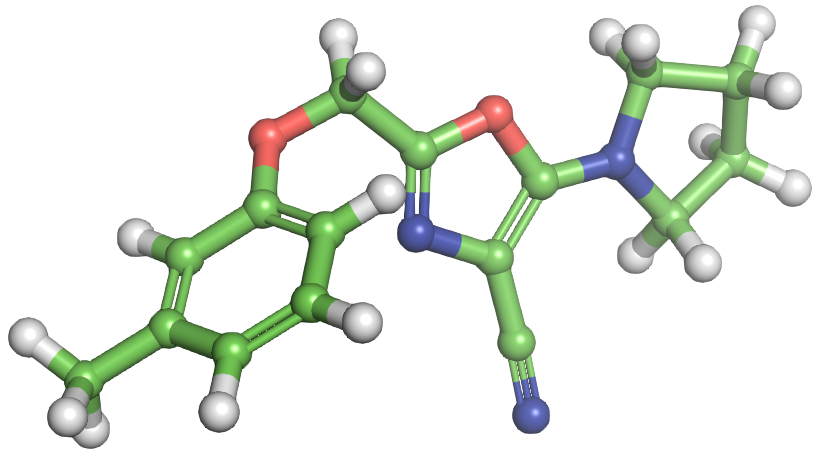} &
\includegraphics[height=1.4cm,width=0.082\linewidth]{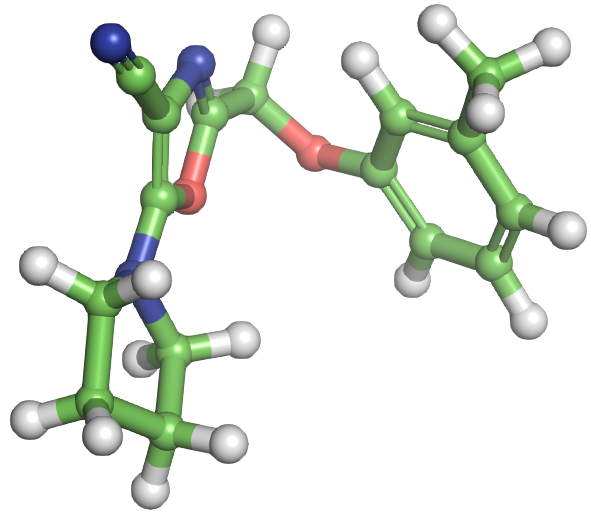} &
\includegraphics[height=1.4cm,width=0.082\linewidth]{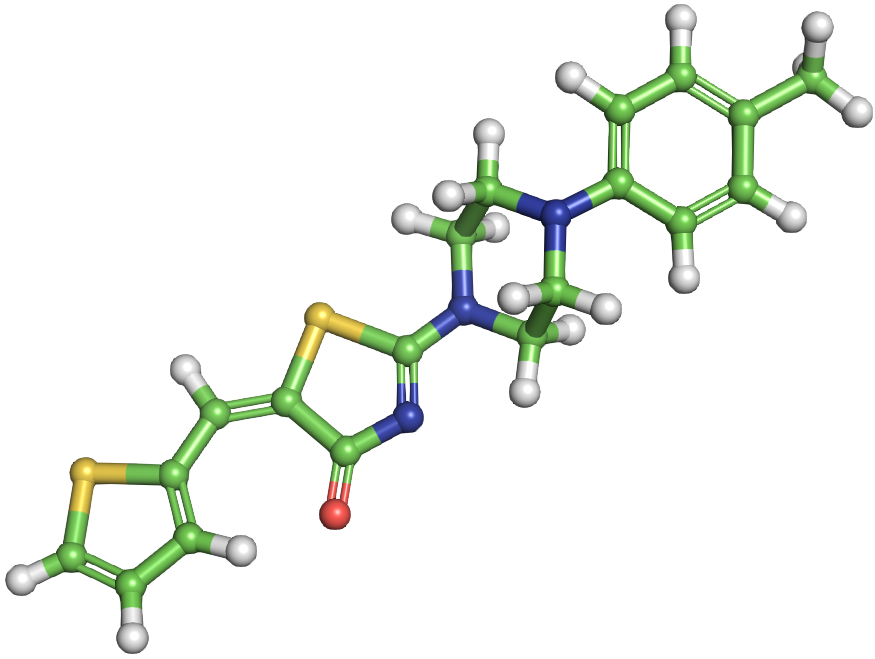} &
\includegraphics[height=1.4cm,width=0.082\linewidth]{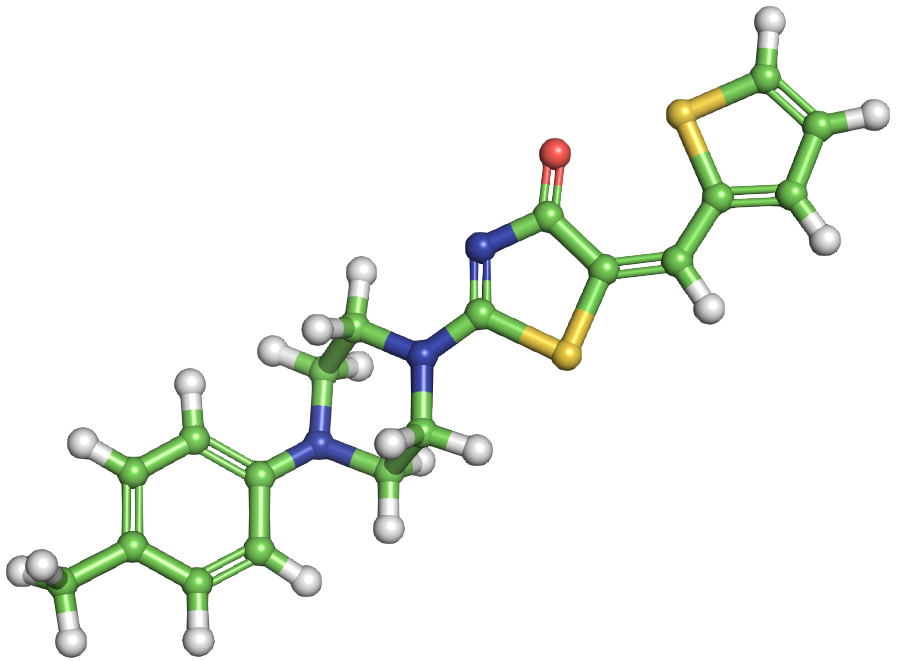} &
\includegraphics[height=1.4cm,width=0.082\linewidth]{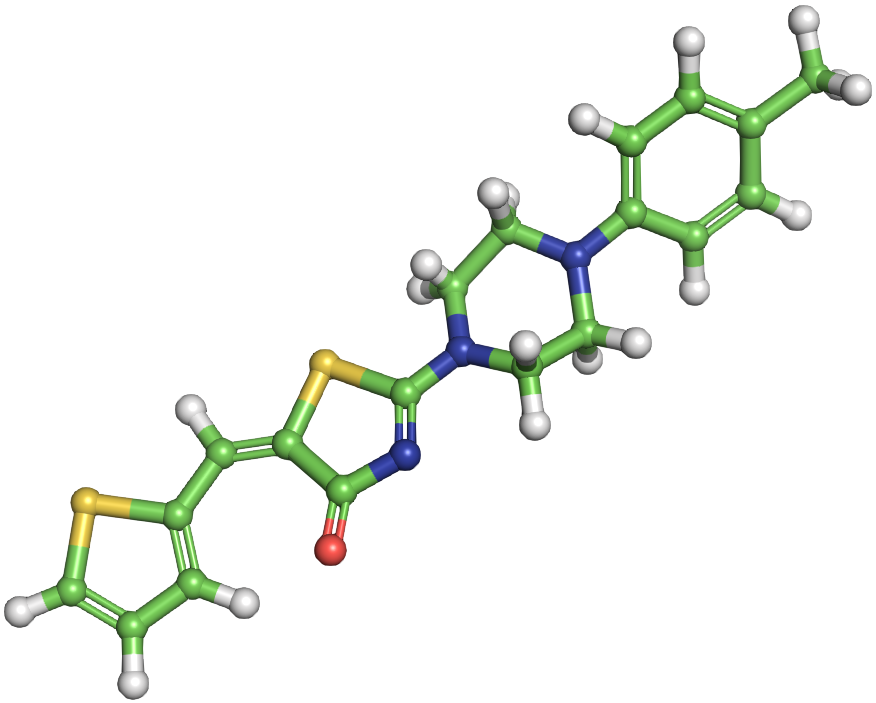}  \\
\rotatebox[origin=l]{90}{ConfGF} &
\includegraphics[height=1.4cm,width=0.082\linewidth]{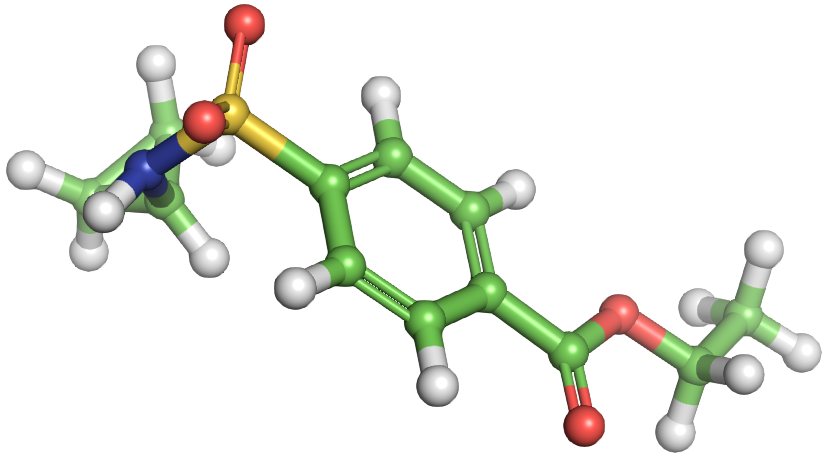} &
\includegraphics[height=1.4cm,width=0.082\linewidth]{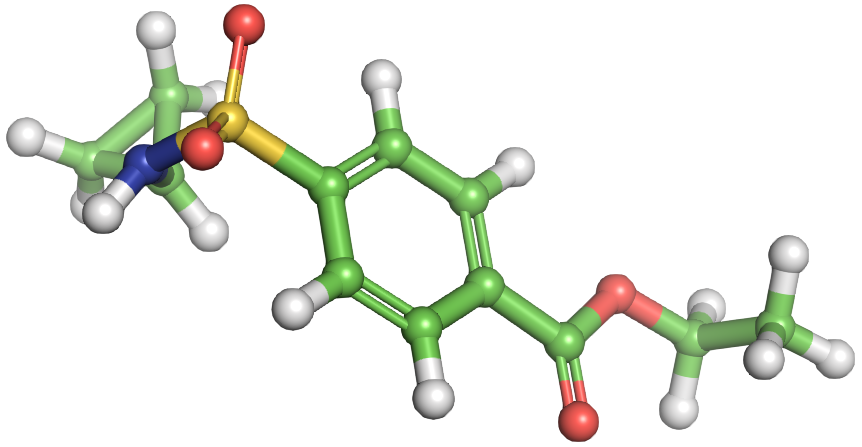} &
\includegraphics[height=1.4cm,width=0.082\linewidth]{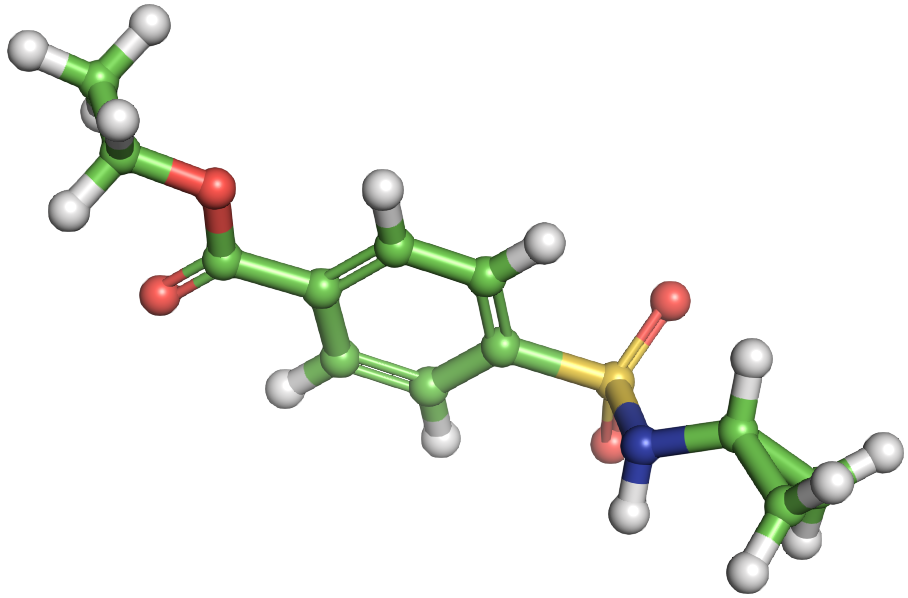} &
\includegraphics[height=1.4cm,width=0.082\linewidth]{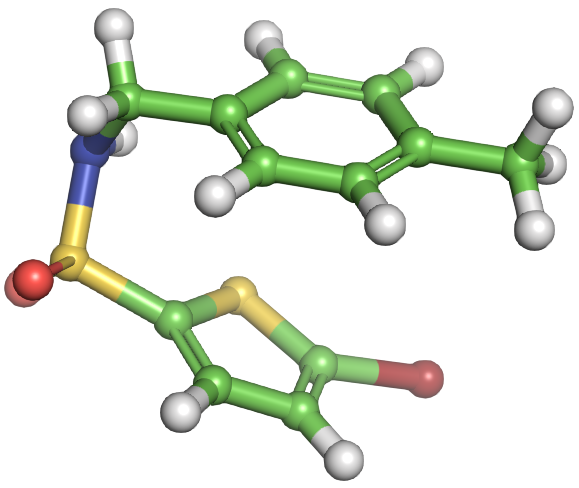} &
\includegraphics[height=1.4cm,width=0.082\linewidth]{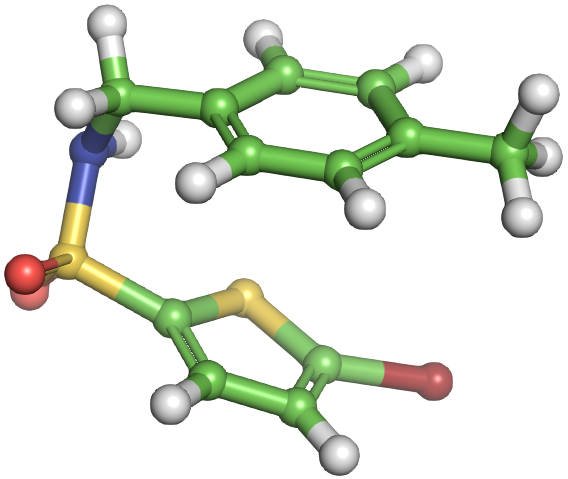} &
\includegraphics[height=1.4cm,width=0.082\linewidth]{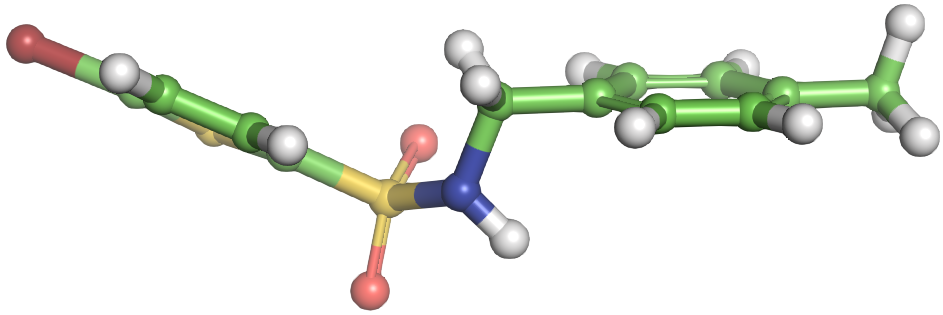} &
\includegraphics[height=1.4cm,width=0.082\linewidth]{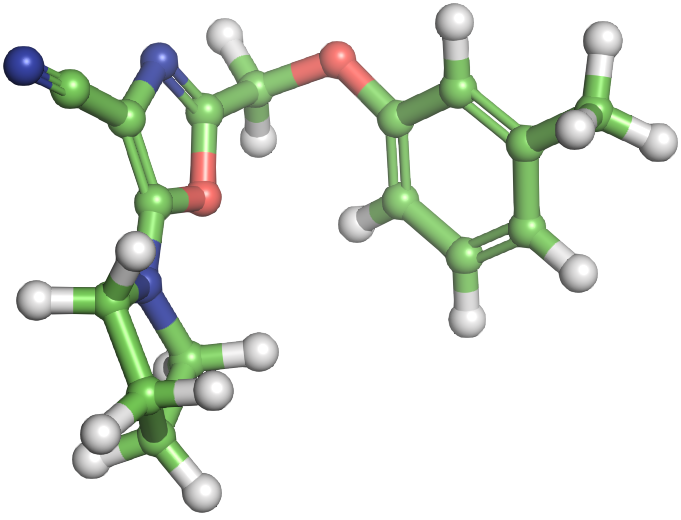} &
\includegraphics[height=1.4cm,width=0.082\linewidth]{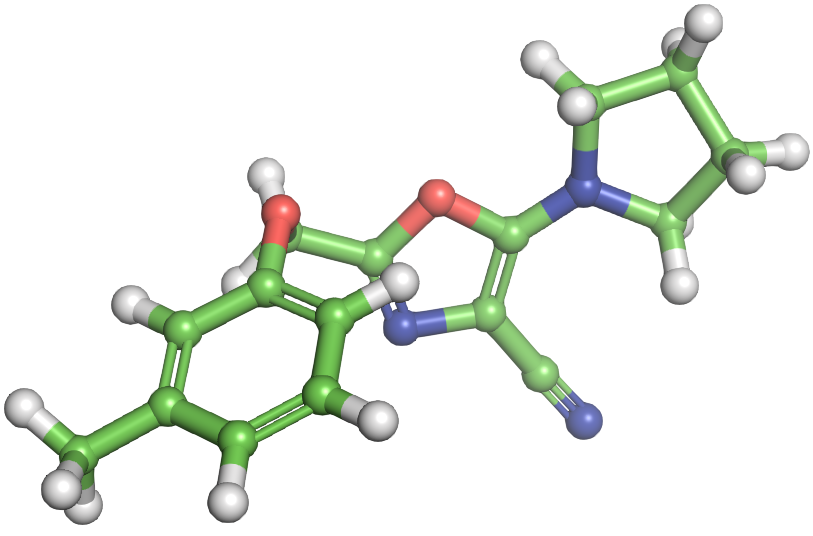} &
\includegraphics[height=1.4cm,width=0.082\linewidth]{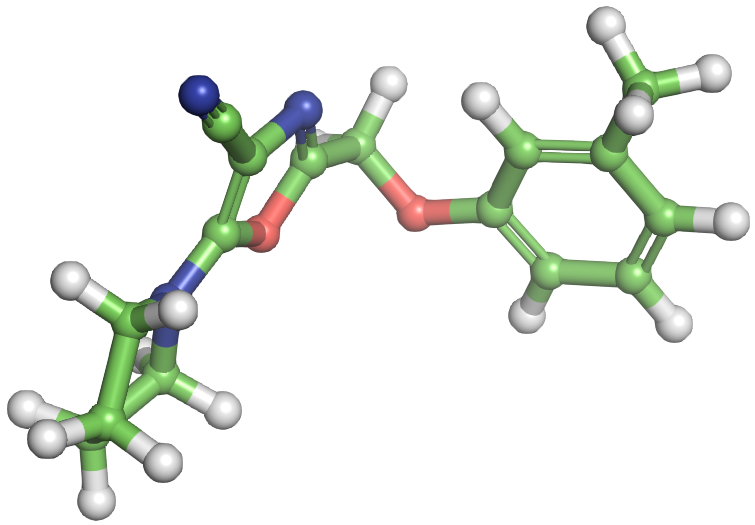} &
\includegraphics[height=1.4cm,width=0.082\linewidth]{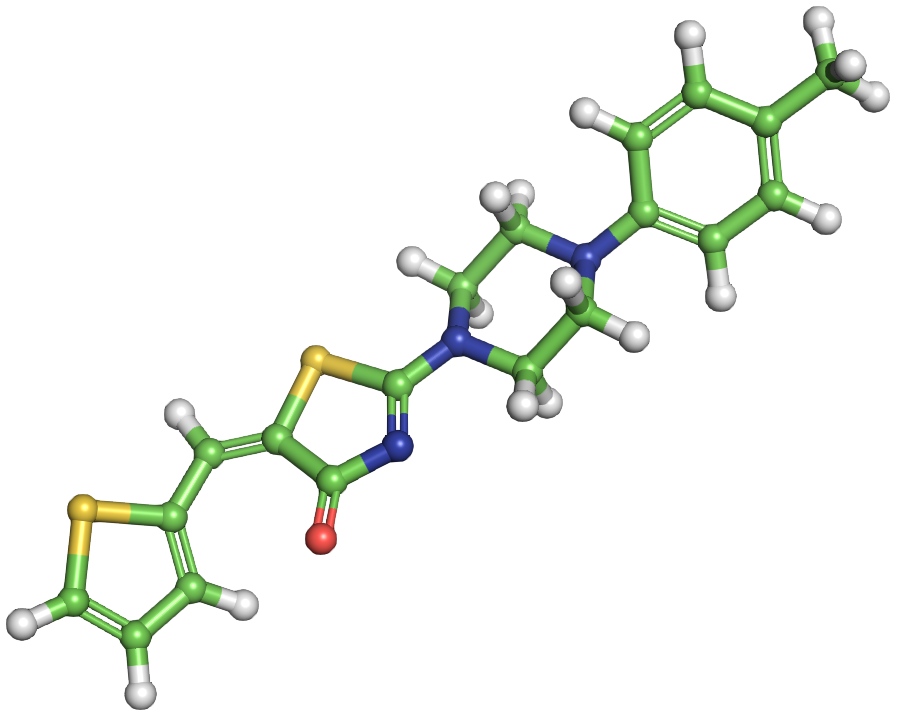} &
\includegraphics[height=1.4cm,width=0.082\linewidth]{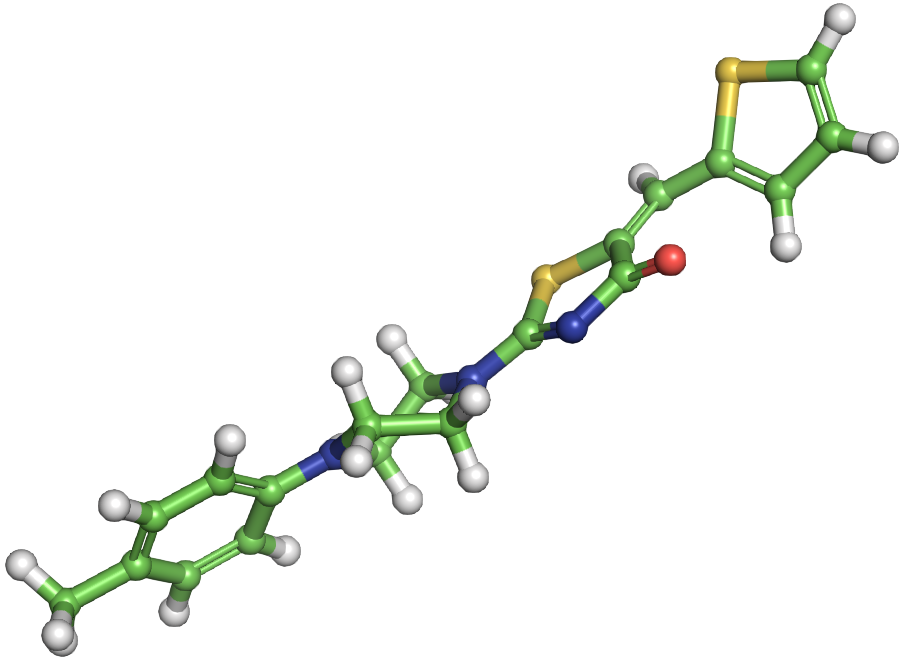} &
\includegraphics[height=1.4cm,width=0.082\linewidth]{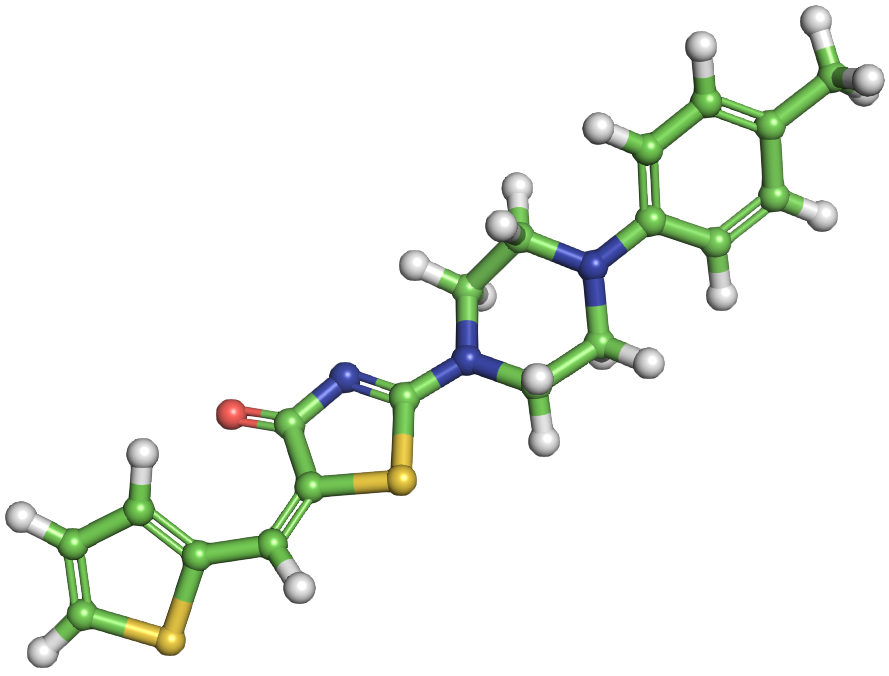}  \\
\rotatebox[origin=l]{90}{CGCF} &
\includegraphics[height=1.4cm,width=0.082\linewidth]{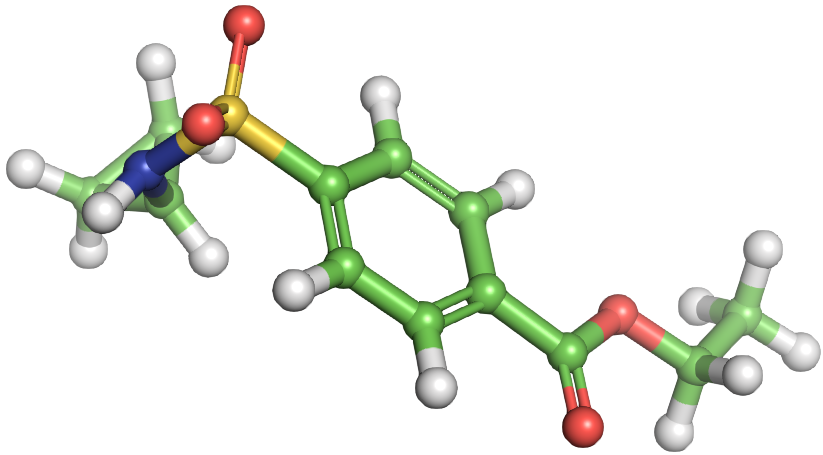} &
\includegraphics[height=1.4cm,width=0.082\linewidth]{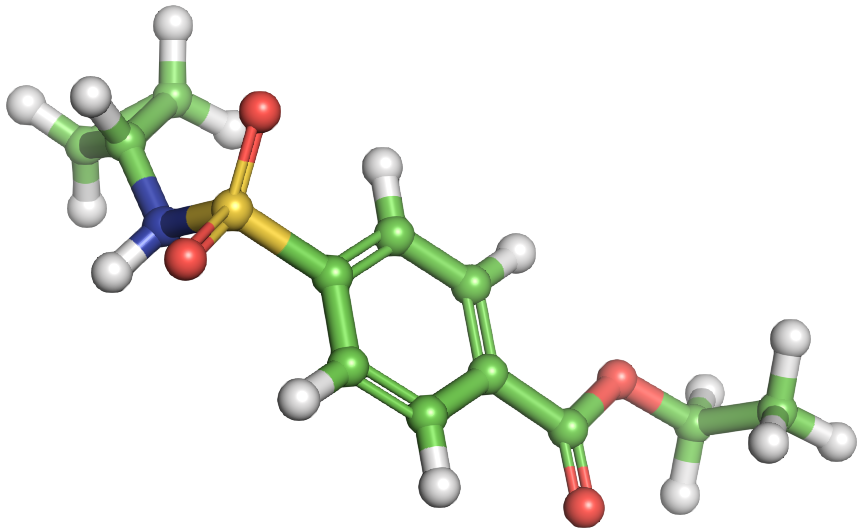} &
\includegraphics[height=1.4cm,width=0.082\linewidth]{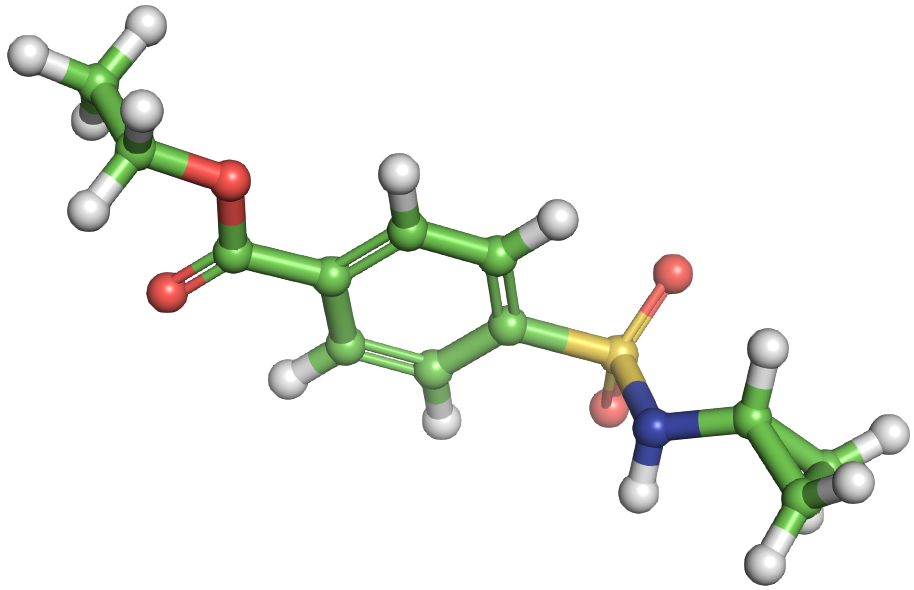} &
\includegraphics[height=1.4cm,width=0.082\linewidth]{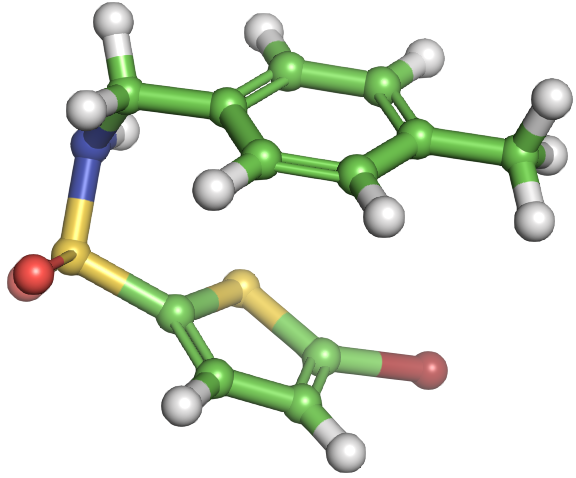} &
\includegraphics[height=1.4cm,width=0.082\linewidth]{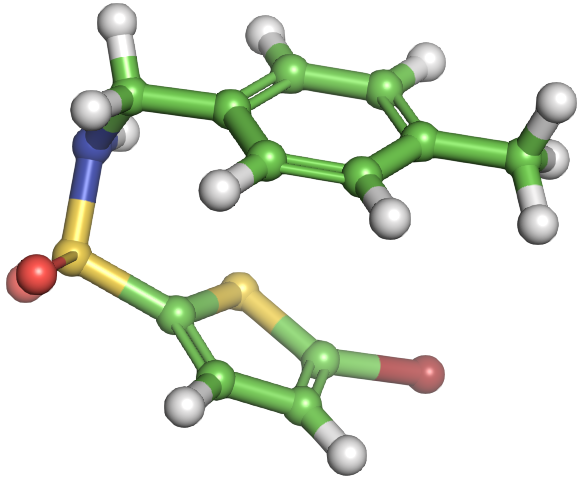} &
\includegraphics[height=1.4cm,width=0.082\linewidth]{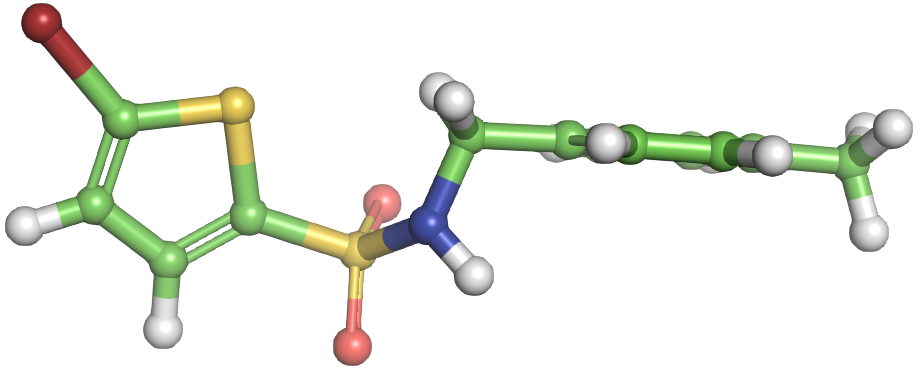} &
\includegraphics[height=1.4cm,width=0.082\linewidth]{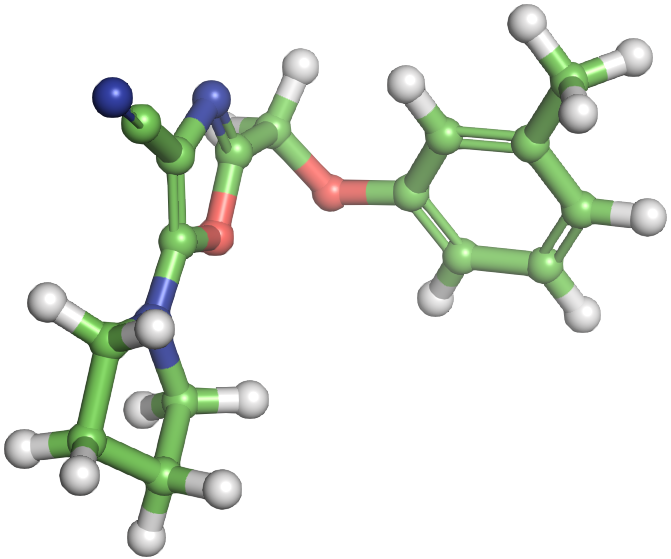} &
\includegraphics[height=1.4cm,width=0.082\linewidth]{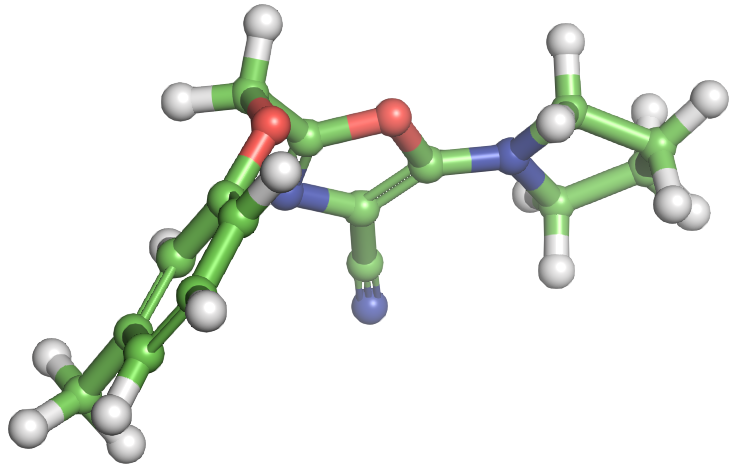} &
\includegraphics[height=1.4cm,width=0.082\linewidth]{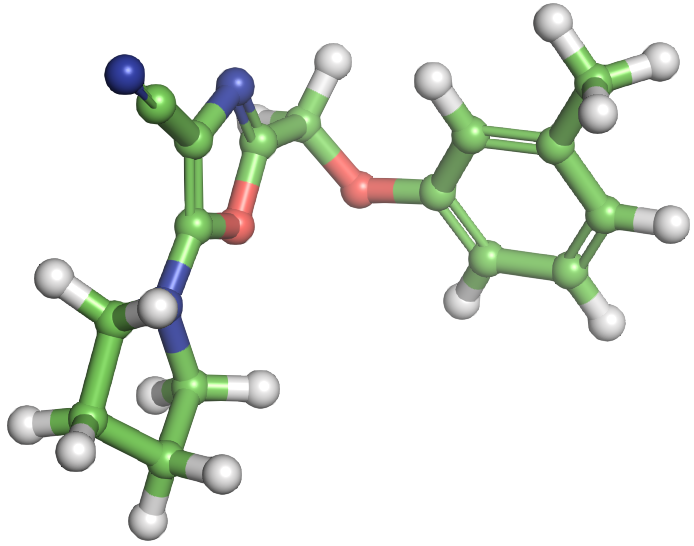} &
\includegraphics[height=1.4cm,width=0.082\linewidth]{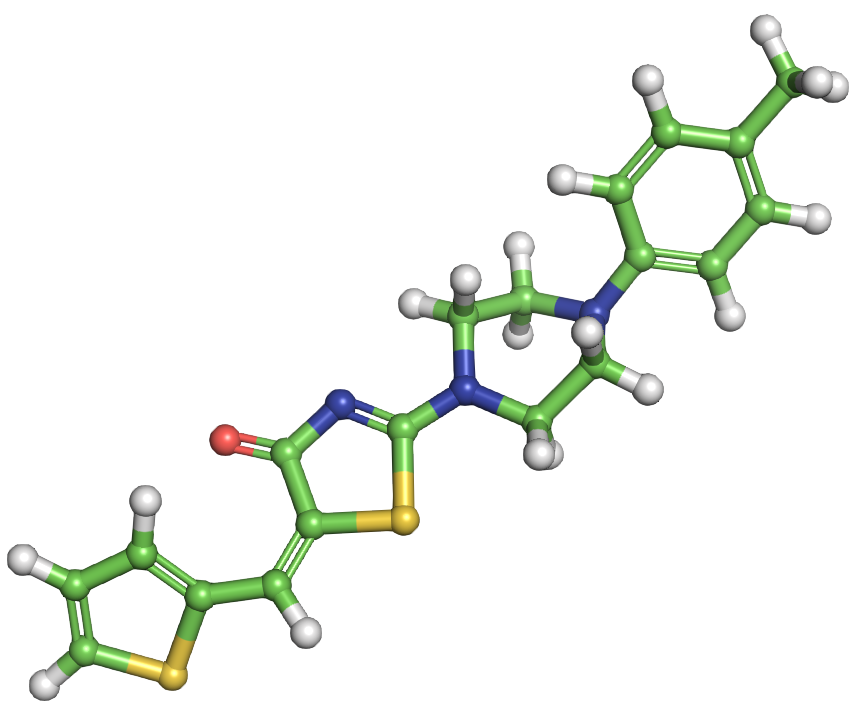} &
\includegraphics[height=1.4cm,width=0.082\linewidth]{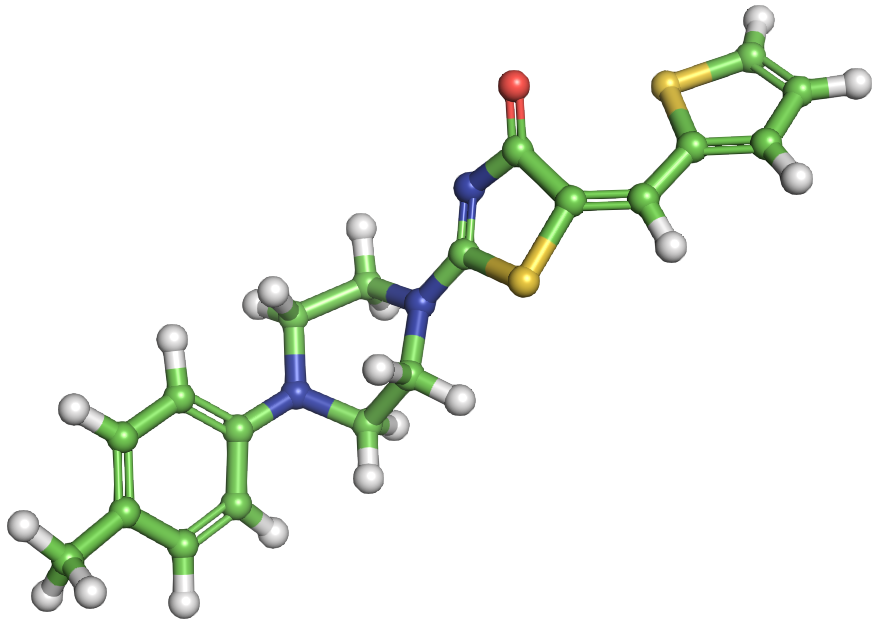} &
\includegraphics[height=1.4cm,width=0.082\linewidth]{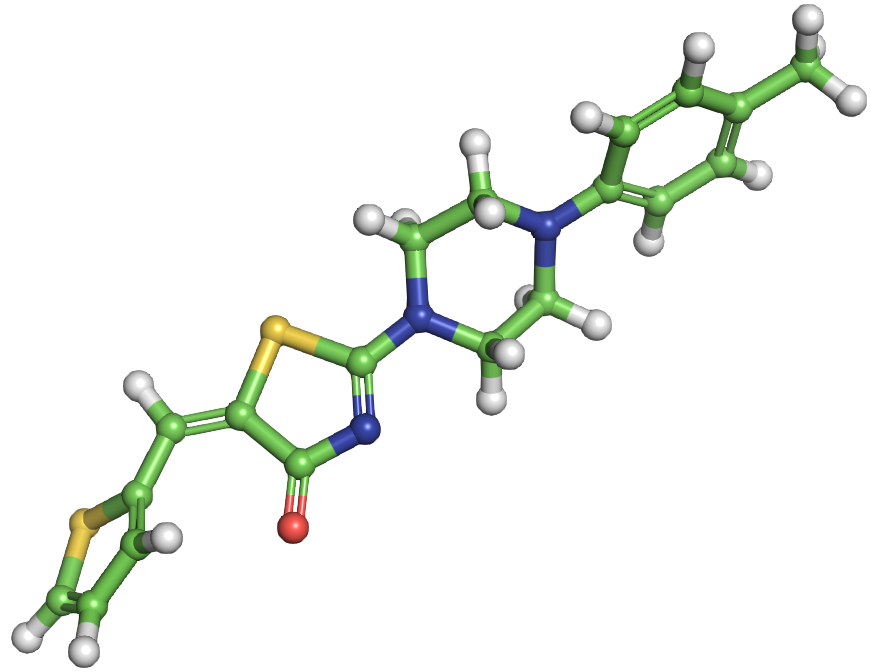}  \\
\bottomrule
\end{tabular}
}
\caption{Visualization of conformations generated by CGCF, ConfGF, and ConfFlow (our approach) for four randomly selected molecular graphs from the test set of GEOM-Drugs. Following a similar protocol to Table~\ref{tab:confgen}, for each approach we sample $2\times$ conformations and pick the one that aligns best with the reference.}
\label{fig:confgenqual}
\end{figure*}

\subsubsection*{Results}
Table \ref{tab:confgen} reports the mean and median of the three scores for all models trained on GEOM-QM9 and GEOM-Drugs. In addition, Table \ref{tab:confgen} also reports the average rank of each algorithm across scores and datasets. For instance, if an algorithm achieves the 2nd best score on half the metrics and the 3rd best on the other half, its average rank is $2.5$.

ConfFlow has the best performance on 6 out of the 12 scores and achieves the best rank overall across datasets and metrics.
In particular, on the large-molecule GEOM-Drugs dataset, ConfFlow outperforms baselines by up to $40\%$.
Note that our model is non-equivariant and directly optimizes for atomic coordinates.
Equivariant models that operate on distances, such as GraphDG, CGCF, ConfVAE, and ConfGF yield lower accuracy because they incur extra error in the conversion from distance geometry to 3D space.
Furthermore, when evaluated by considering all the atoms (i.e., including H), the coverage score attained by the baselines drops dramatically, as can be seen in Table \ref{tab:confgenH}.
The ConfFlow performance drops the least and is $3\times$ better than ConfGF, the next best learning-based method.
ConfFlow is the only neural model that consistently outperforms the rule-based RDKit on COV and MAT scores across datasets without relying on any extra force-field computation \cite{xu2021end,xu2021learning}.
Figure~\ref{fig:confgenqual} illustrates conformations sampled by trained CGCF, ConfGF, and ConfFlow models. ConfFlow generates more realistic and diverse structures.
\begin{table}[!tbhp]
\centering
\ra{1.05}
\resizebox{0.55\linewidth}{!}{
\begin{tabular}{@{}l@{\hspace{1mm}}|c@{\hspace{2mm}}c@{\hspace{2mm}}c@{\hspace{2mm}}c@{\hspace{2mm}}c@{\hspace{2mm}}c@{}}
\toprule
\multirow{2}{*}{Algorithm} & \multicolumn{2}{c}{COV ($\%$)} & \multicolumn{2}{c}{MAT ($\r{A}$)} & \multicolumn{2}{c}{MIS ($\%$)} \\
& Mean & Median & Mean & Median & Mean & Median \\
\midrule
CGCF & 6.90 & 0.50 & 1.782 & 1.753 & 97.30 & 99.70 \\
ConfGF & 9.00 & 2.20 & 1.717 & 1.690 & 96.70 & 99.30 \\
\textbf{ConfFlow} & \textbf{31.60} & \textbf{25.50} & \textbf{1.427} & \textbf{1.380} & \underline{84.30} & \underline{90.00} \\
\midrule
RDKit & \underline{28.20} & \underline{22.60} & \underline{1.678} & \underline{1.578} & \textbf{74.20} & \textbf{84.00} \\
\bottomrule
\end{tabular}
}
\caption{Results on GEOM-Drugs when measured using RMSD over all atoms, including H.}
\label{tab:confgenH}
\end{table}

\subsection{Property Prediction}
\begin{table}[!tbhp]
\centering
\ra{1.05}
\resizebox{0.6\linewidth}{!}{
\begin{tabular}{@{}l@{\hspace{1mm}}|c@{\hspace{2mm}}c@{\hspace{2mm}}c@{\hspace{2mm}}c@{\hspace{2mm}}c@{\hspace{2mm}}c@{}}
\toprule
Algorithm & $E_{\text{min}}$ & $\Delta\epsilon_{\text{min}}$ & $\Delta\epsilon_{\text{max}}$ & $\epsilon_{\text{HOMO}}$ & $\epsilon_{\text{LUMO}}$ & $\mu$ \\
\midrule
GraphDG & 25.9901	& 2.8715 & 5.8113 & 3.3564 & 1.1126 & 2.4256 \\
CGCF & 15.0759 & 1.0777 & 7.0687	& 1.9896 & 1.7726 & 2.4816 \\
ConfGF & 2.7526 & 0.2197 & 5.5104 & 1.7925 & 0.8149 & 1.3028 \\
\textbf{ConfFlow} & \textbf{0.9927} & \textbf{0.1801} & \underline{5.0715} & \underline{0.9380} & \textbf{0.1280} & \textbf{0.8962} \\
\midrule
RDKit & \underline{1.8400} & \underline{0.1953} & \textbf{0.3515} & \textbf{0.1164} & \underline{0.2665} & \underline{1.1098} \\
\bottomrule
\end{tabular}
}
\caption{Median of absolute prediction errors of various ensemble properties measured on GEOM-Drugs. Properties related to energy are reported in eV and the dipole moment $\mu$ in debye.}
\label{tab:proppred}
\end{table}
This task measures ensemble properties of a molecular graph, each calculated by aggregating the property estimate from multiple conformations. We consider 30 molecular graphs randomly drawn from the test set of GEOM-Drugs. For each conformation of a molecular graph we first compute the total energy, highest occupied molecular orbital (HOMO) and lowest unoccupied molecular orbital (LUMO) energies, and the dipole moment using the quantum chemical calculation package \textit{Psi4} \cite{turney2012psi4}. Then the ensemble properties -- lowest energy $E_{\text{min}}$, mean energies of HOMO and LUMO, minimum and maximum HOMO-LUMO gap $\Delta\epsilon$, and average dipole moment $\mu$ -- of each molecular graph are calculated based on their corresponding conformational properties. Following this, we sample 50 conformations per molecular graph from each baseline and compute ensemble properties by repeating the above mentioned procedure. 

\subsubsection*{Results}
We use the median of absolute error (MedAE) w.r.t. the test set to measure the accuracy of predicted ensemble properties~\cite{Simm2020GraphDG, shi*2021confgf}. As reported in Table \ref{tab:proppred}, ConfFlow is the only learning-based method that achieves lower MedAE than RDKit in four of the six properties; it is the next best on the other two.
The better performance of RDKit on $\Delta\epsilon_{\text{max}}$ (maximum HOMO-LUMO gap) is due to the presence of occasional outliers in the conformations synthesized by learning-based methods. A large error for $\Delta\epsilon_{\text{max}}$ can be induced by even a small number of outlier conformers.

\section{Conclusion}
We have presented a new computational approach to generating 3D conformations of any molecular graph. Our approach, ConfFlow, is based on continuous normalizing flows and does not enforce any explicit geometric constraints. This enables ConfFlow to quickly generate stable and diverse structures, yielding high accuracy in conformation generation and property prediction of large drug-like molecules.

\bibliography{paper}
\bibliographystyle{paper}

\appendix
\section{Additional Experiments}
\subsection{Distribution over Distances}
In \cite{Simm2020GraphDG}, authors considers additional task that evaluates whether the method can model the underlying distance distribution. This is estimated by computing  maximum mean discrepancy (MMD) of generated distances relative to the ground-truth distance distribution extracted from reference conformations. This metric assesses how closely the generated distribution over distances matches that of the ground truth. As ConfFlow do not explicitly model molecular distance geometry, we substitute interatomic distances calculated from the generated coordinates. For each molecular graph $\G$ in the test set, we sample the same number of conformations as found in the test set. Following the evaluation procedure recommended by \cite{Simm2020GraphDG, xu2021learning}, we ignore edges associated with H atoms and evaluates MMD between the generated and ground-truth distributions for three different statistics -- the joint distribution over all distances $p(\dd | \G)$ (All), the pairwise marginal distribution $p(d_{ij}, d_{uv} | \G)$ (Pair), and the marginal distribution of individual distances $p(d_{ij} | \G)$ (Single).

\subsubsection*{Results}
We aggregate results over all graphs in the test set and report mean and median MMD in Table \ref{tab:dist} on GEOM-QM9 dataset. We observe that the distance distribution of baselines that explicitly optimize for molecular distance geometry is significantly closer to the ground-truth distribution than the models that train only on atomic coordinates. In particular, ConfVAE and CGCF yield the lowest MMD despite performing moderately on the conformation generation task. This is due to the bilevel optimization setup in ConfVAE, which tightly grounds distances using multiple losses. On the other hand, ConfFlow falls short on the distance modeling task, with its MMD on par with ConfGF. In contrast, other coordinate-based models such as CVGAE and RDKit seem to struggle with the distance modeling task. Since RDKit is designed to generate conformations only at equilibrium states, i.e., distribution modes, it fails to model the overall underlying distribution. Furthermore, for GraphDG and ConfFlow model the MMD score improves marginally on incorporating the distances w.r.t. hydrogen atoms.

\subsection{Ablation Studies}
We conduct a number of ablation studies on the GEOM-QM9 dataset. Table \ref{tab:ablation} shows the accuracy of various ablated forms of our model. We begin by varying the number of point transformer blocks $S$ within GPTF. We observe that removing or adding a point transformer block only marginally effects the overall accuracy. For the rest of the ablation study we fix $S=2$. This is done so that all ablated models can be trained with the same batch size per GPU, factoring out variation in other hyperparameters on the results.

Next we experiment with varying the depth $L$ of ConfFlow and the number $R$ of message passing layers. Here, $L$ denotes total number of reversible batch normalization (BN) and GCF blocks in ConfFlow. By default we set $R=2$ and $L=5$, that consists of three BN layers interleaved with two blocks of GCF. If we reduce each of these settings by a factor of two, performance declines. Doubling of these settings yields only a marginal increase in accuracy. In general, we observed that the performance of ConfFlow is most affected by the variation in the basic processing block of the model, i.e., the message passing layer. For instance, replacing all point transformer message passing layers with GN layers \cite{battaglia2018relational} leads to overall performance degradation. We also tried replacing GCFs with recently introduced reversible GNN layer named GRevNet~\cite{liu2019gnf}. As with discrete normalizing flows, the invertibility in GRevNets is achieved by splitting and alternatingly updating each part of 3D coordinates while conditioned on the other parts. However, for molecular setting we find GRevNets training to be quite unstable with model collapsing within few iterations.

Finally, we quantify the effect of training regularization terms. As expected, disabling regularization reduces test accuracy. Simply regularizing the Jacobian Frobenius norm achieves stable training and comparable performance to ConfFlow.

\begin{table}
\centering
\begin{tabular}{@{}l@{\hspace{1mm}}|c@{\hspace{2mm}}c@{\hspace{2mm}}c@{\hspace{2mm}}c@{\hspace{2mm}}c@{\hspace{2mm}}c@{\hspace{2mm}}|c@{\hspace{2mm}}c@{\hspace{2mm}}c@{\hspace{2mm}}c@{\hspace{2mm}}c@{\hspace{2mm}}c@{}}
\toprule
Dataset & \multicolumn{6}{c|}{No Hydrogen} & \multicolumn{6}{c}{With Hydrogen} \\
\midrule
\multirow{2}{*}{Algorithm} & \multicolumn{2}{c}{Single} & \multicolumn{2}{c}{Pair} & \multicolumn{2}{c|}{All} & \multicolumn{2}{c}{Single} & \multicolumn{2}{c}{Pair} & \multicolumn{2}{c}{All} \\
& Mean & Median & Mean & Median & Mean & Median & Mean & Median & Mean & Median & Mean & Median \\
\midrule
CVGAE & 6.710 & 7.208 & 6.976 & 7.286	& 7.074 &	7.230 & 6.851 & 7.270 & 7.031 & 7.243 & 7.073 & 7.216 \\
GraphDG & 3.628 & 3.631 & 4.120 & 4.113 & 4.571 & 4.491 & 2.033 & 1.787 & 2.051 & 1.874 & 2.030 & 1.987 \\
CGCF & \underline{0.250} & \underline{0.161} & \underline{0.237} & \underline{0.163} & \underline{0.216} & \textbf{0.169} & \underline{0.260} & \underline{0.151} & \underline{0.233} & \textbf{0.148} & \underline{0.214} & \textbf{0.160} \\
ConfVAE & \textbf{0.209} & \textbf{0.151} & \textbf{0.204} & \textbf{0.158} & \textbf{0.198} & \textbf{0.169} & \textbf{0.190} & \textbf{0.146} & \textbf{0.177} & \textbf{0.148} & \textbf{0.177} & \underline{0.162} \\
ConfGF & 0.951 & 0.846 & 1.111 & 1.039 & 1.303 & 1.240 & 1.307 & 1.319 & 1.539 & 1.518 & 1.939 & 1.933 \\
\textbf{ConfFlow} & 1.454 & 1.412 & 1.735 & 1.739 & 2.023 & 2.013 & 1.351 & 1.311 & 1.586 & 1.580 & 2.034 & 2.040 \\
\midrule
RDKit & 3.378 & 3.335 & 3.762 & 3.790 & 4.107 & 4.241 & 3.518 & 3.611 & 3.934 & 3.991 & 4.188 & 4.191 \\
\bottomrule
\end{tabular}
\caption{Comparison of mean and median MMD between the distance distribution of ground-truth and generated conformations on the GEOM-QM9 datasets.}
\label{tab:dist}
\end{table}
\begin{table}
\centering
\begin{tabular}{@{}l@{\hspace{1mm}}|c@{\hspace{2mm}}|c@{\hspace{2mm}}c@{\hspace{2mm}}c@{\hspace{2mm}}c@{\hspace{2mm}}c@{\hspace{2mm}}c@{}}
\toprule
\multirow{2}{*}{Ablate} & \multirow{2}{*}{Params} & \multicolumn{2}{c}{COV ($\%$)} & \multicolumn{2}{c}{MAT ($\r{A}$)} & \multicolumn{2}{c}{MIS ($\%$)} \\
& & Mean & Median & Mean & Median & Mean & Median \\
\midrule
- & 1.7M & 91.08 & \textbf{95.60} & 0.278 & 0.287 & 0.476 & 0.507 \\
\midrule
$S=2$ & 1.1M & 90.20 & 94.20 & 0.300 & 0.307 & 0.490 & 0.525 \\
$S=4$ & 2.3M & \underline{91.90} & \textbf{95.60} & \underline{0.263} & \underline{0.269} & \underline{0.455} & \textbf{0.480} \\
\midrule
$L=3$ & 0.6M & 80.50 & 85.60 & 0.359 & 0.373 & 0.544 & 0.588 \\
$L=9$ & 2.2M & \textbf{92.00} & 94.80 & 0.294 & 0.299 & 0.477 & 0.524 \\
\midrule
$R=1$ & 0.7M & 75.70 & 77.10 & 0.398 & 0.405 & 0.572 & 0.616 \\
$R=4$ & 2.0M & \underline{91.90} & \underline{95.40} & \textbf{0.253} & \textbf{0.252} & \textbf{0.436} & \underline{0.483} \\
\midrule
GN & 1.4M & 86.70 & 92.40 & 0.347 & 0.352 & 0.535 & 0.573 \\
\midrule
w/o regularization & 1.1M & 88.40 & 92.60 & 0.305 & 0.311 & 0.490 & 0.530 \\
only KE & 1.1M & 55.30 & 55.30 & 0.484 & 0.481 & 0.813 & 0.848 \\
only JN & 1.1M & 90.00 & 94.80 & 0.294 & 0.296 & 0.479 & 0.514 \\
\bottomrule
\end{tabular}
\caption{Ablation experiments on the GEOM-QM9 dataset.}
\label{tab:ablation}
\end{table}

\section{Input Features}
Table~\ref{tab:node} and Table~\ref{tab:edge} lists node and edge attributes used across all experiments and tasks.
\begin{table}
\centering
\begin{tabular}{@{}l@{\hspace{2mm}}l@{\hspace{2mm}}l@{}}
\toprule
Features & Data Type & Dimension \\
\midrule
atomic number & integer \{1-119, None\} & 1 \\
hybridization & integer \{'S', 'SP', 'SP2', 'SP3', 'SP3D', 'SP3D2', None\} & 1 \\
atomic degree & integer \{0-10, None\}  & 1 \\
formal charge & integer \{-5 - +5, None\} & 1 \\
total hydrogen atoms & integer \{0-8, None\} & 1 \\
implicit valence electrons & integer \{ 1-15, None \} & 1 \\
total valence electrons & integer \{ 1-15, None \} & 1 \\
total radical electrons & integer \{0-4, None \} & 1 \\
chirality & integer \{ 'CW', 'CCW', Others, None\} & 1 \\
is aromatic & binary & 1 \\
is in ring & binary & 1 \\
\bottomrule
\end{tabular}
\caption{Node Attributes.}
\label{tab:node}
\end{table}
\begin{table}
\centering
\begin{tabular}{@{}l@{\hspace{2mm}}l@{\hspace{2mm}}l@{}}
\toprule
Features & Data Type & Dimension \\
\midrule
type & integer \{'single', 'double', 'triple', 'aromatic', None \} & 1 \\
stereo chemistry & integer \{'Z', 'E', 'CIS', 'TRANS', 'ANY', None\} & 1 \\
is conjugated & binary & 1 \\
is same ring & binary & 1 \\
shortest path & integer \{ 1-3, None\} & 1 \\
is in ring of size (3-9) & binary & 7 \\
\bottomrule
\end{tabular}
\caption{Edge Attributes.}
\label{tab:edge}
\end{table}

\section{Supplement video}
It demonstrates the step-by-step transformation of molecular conformation, starting from initial Gaussian noise and progressing to the final 3D equilibrium structure through ConfFlow.

\end{document}

%% file: vcl-shortcuts.tex
\usepackage{graphicx}
\usepackage[caption=false]{subfig}
\usepackage{amsmath}
\usepackage{amssymb}
\usepackage{epstopdf}
\usepackage{bm,xspace}
\usepackage{verbatim}
\usepackage{multirow}
\usepackage{wrapfig}
\usepackage{balance}
\usepackage{booktabs}
\usepackage{calc}

\newcommand{\ra}[1]{\renewcommand{\arraystretch}{#1}}

\def\dd{{\mathbf{d}}}

\def\hh{{\mathbf{h}}}

\def\mm{{\mathbf{m}}}

\def\xx{{\mathbf{x}}}

\def\zz{{\mathbf{z}}}

\def\AA{{\mathbf{A}}}

\def\HH{{\mathbf{H}}}

\def\RR{{\mathbf{R}}}

\def\XX{{\mathbf{X}}}
\def\YY{{\mathbf{Y}}}
\def\ZZ{{\mathbf{Z}}}

\def\A{{\mathcal{A}}}

\def\E{{\mathcal{E}}}
\def\F{{\mathcal{F}}}
\def\G{{\mathcal{G}}}

\def\M{{\mathcal{M}}}
\def\N{{\mathcal{N}}}

\def\S{{\mathcal{S}}}

\def\V{{\mathcal{V}}}

\def\Re{{\mathbb{R}}}
\def\Se{{\mathbb{S}}}

\def\btheta{{\bm\theta}}

\DeclareMathOperator*{\argmax}{arg\,max}

\DeclareMathOperator{\trace}{tr}

\newcommand{\paren}[1]{\left(#1\right)}
\newcommand{\abs}[1]{\left\lvert#1\right\rvert}
\newcommand{\norm}[1]{\left\lVert#1\right\rVert}

\newcommand{\timess}{\mathbin{\!\times\!}}